\documentclass[a4paper,fleqn]{cas-dc}

\usepackage[numbers]{natbib}
\usepackage{amsmath,amssymb,amsfonts}
\usepackage{algorithm,algorithmic}
\usepackage{graphicx}
\usepackage{textcomp}
\usepackage{makecell}
\usepackage{adjustbox}
\usepackage{xcolor}
\usepackage{tikz}
\usepackage{pgfplots}
\usepackage{subfig}
\usepackage{subcaption}
\usepackage{multicol}
\usepackage{multirow}
\usepackage{pgfplots}
\usepackage{balance}
\pgfplotsset{compat=1.18}
\usepackage{balance}

\def\tsc#1{\csdef{#1}{\textsc{\lowercase{#1}}\xspace}}
\tsc{WGM}
\tsc{QE}
\tsc{EP}
\tsc{PMS}
\tsc{BEC}
\tsc{DE}

\begin{document}
\let\WriteBookmarks\relax
\def\floatpagepagefraction{1}
\def\textpagefraction{.001}

\shorttitle{SynPre-FL: Synthetic data-driven pretraining integrated FL training framework}

\shortauthors{A.K.Nair et~al.}


\title [mode = title]{SynPre-FL: Synthetic data-driven pretraining integrated Federated Learning training framework}



\tnotetext[1]{The author(s) declare that financial support was received for the research and/or publication of this article. This study was supported by UKRI through the Horizon Europe Guarantee Scheme (project number: 10078953) for the European Commission-funded PHASE IV AI project (grant agreement number: 101095384) under the Horizon Europe Programme.} 


%
\author[1]{Akarsh K Nair}[orcid=0000-0002-7734-0367]

\cormark[1]
\ead{akarsh.kongasserynair@ntu.ac.uk}

\credit{Conceptualisation of this study, Methodology, Software, Implementation, Paper Writing, Review}
\affiliation[1]{organization={Department of Computer Science},
    addressline={Nottingham Trent University},  city={Nottingham},
    country={United Kingdom}}

\author[1]{Muhammad Arifur Rahman}[orcid=0000-0002-6774-0041]
\ead{arif.rahman@ntu.ac.uk}

\credit{Conceptualization of this study, Methodology, Software}

\author[1]{Nicholas Shopland}[orcid=0000-0003-2082-9070]
\ead{nicholas.shopland@ntu.ac.uk}

\credit{Conceptualization of this study, Methodology, Software}

\author[1]{Andy Burton}[orcid=0000-0002-9073-8310]
\ead{andrew.burton@ntu.ac.uk}

\credit{Conceptualization of this study, Methodology, Software}

\author[1]{Jun He}[orcid=0000-0002-5616-4691]
\ead{jun.he@ntu.ac.uk}

\credit{Conceptualization of this study, Methodology, Software}


\author[1]{Yuan Shen}[]
\ead{yuan.shen@ntu.ac.uk}
\author[2]{David Baldwin}[]
\affiliation[2]{organization={Division of Epidemiology and Public Health},
    addressline={University of Nottingham}, 
    city={Nottingham},
    country={United Kingdom}}
\ead{david.baldwin@nottingham.ac.uk}
\author[2]{Emma O'Dowd}[]
\ead{emma.odowd@nottingham.ac.uk}
\author[2]{Amna Burzic}[]
\ead{amna.burzic@nottingham.ac.uk}

\credit{Conceptualization of this study, Methodology, Software}

\author[3,4]{Mufti Mahmud}[orcid=0000-0002-2037-8348]
\affiliation[3]{organization={Information and Computer Science Department},
    addressline={King Fahd University of Petroleum and Minerals}, 
    city={Dhahran},
    country={Saudi Arabia}}
\affiliation[4]{organization={SDAIA-KFUPM Joint Research Center for AI},
addressline={King Fahd University of Petroleum and Minerals}, 
city={Dhahran},
country={Saudi Arabia}}
    \ead{mufti.mahmud@kfupm.edu.sa}

\credit{Conceptualization of this study, Methodology, Software}

\author[1]{David J. Brown}[orcid=0000-0002-1677-7485]
\ead{david.brown@ntu.ac.uk}

\credit{Conceptualisation of this study, Methodology, Software}

\cortext[cor1]{Corresponding author}



\begin{abstract}
Federated learning (FL) has emerged as a promising paradigm for privacy-preserving clinical risk prediction, yet its practical deployment in healthcare remains constrained by limited data shareability, strong client heterogeneity, class imbalance, and the lack of realistic benchmark datasets for tabular electronic health records (EHRs). In parallel, while synthetic data generation offers a potential solution to data scarcity, its integration with federated optimisation pipelines has not been systematically studied.
In this work, we propose SynPre-FL, a unified framework that combines high-fidelity synthetic EHR generation with synthetic-pretrained FL for robust clinical prediction in non-IID settings. The framework employs a latent autoencoder–diffusion model to generate privacy-preserving synthetic EHR cohorts, which are used to warm-start federated training through a synthetic pretraining phase. This initialisation is followed by heterogeneity-aware federated optimisation incorporating class-balanced local objectives, proximal regularisation, and adaptive server-side aggregation. Post-hoc probability calibration and federated-safe explainability are integrated to ensure reliable and interpretable clinical risk estimates.
Extensive experiments demonstrate that the proposed synthetic generator preserves the univariate, bivariate, and multivariate statistical structure while providing strong privacy protection against membership inference and reconstruction attacks. Synthetic data exhibit high downstream utility under TSTR/TRTS protocols and model-based evaluations. In federated experiments with 5, 10, and 15 heterogeneous clients, SynPre-FL consistently improves robustness and scalability compared to 
baselines, particularly under severe non-IID fragmentation. Calibration further enhances the reliability of probability estimates, and SHAP-based explainability reveals stable and clinically coherent feature attributions across federation sizes. In overview, SynPre-FL provides a practical and reproducible framework for integrating synthetic data generation with FL, enabling privacy-aware, interpretable, and robust clinical prediction from distributed tabular EHR data.
\end{abstract}


\begin{highlights}

\item Hybrid autoencoder–diffusion model for high-fidelity synthetic EHR generation
\item A synthetic-pretraining integrated training framework (SynPre-FL) for improving federated learning under non-IID settings
\item Comprehensive privacy and utility evaluation for synthetic clinical data
\item Federated-safe explainability framework for feature level contribution interpretation without data exposure
\item Reproducible benchmark for clinical tabular federated learning
\end{highlights}


\maketitle


\section{Introduction}

Artificial intelligence (AI) for healthcare care is increasingly based on large-scale, high-dimensional electronic health record (EHR) data to
enable accurate risk prediction, early disease detection, and robust
clinical decision support. However, the development of such models remains severely constrained due to privacy regulations~\cite{PATI2024100974,TEO2024101419}, restricted data sharing between institutions, and the scarcity of publicly available benchmark datasets for tabular clinical prediction tasks. These limitations hinder the reproducibility of the research, slow methodological innovation, and prevent fair and consistent comparisons between algorithms. As a result, synthetic data generation has emerged as a promising alternative, offering the potential to reproduce statistical properties of real EHR populations~\cite{giuffre2023harnessing} while eliminating patient-identifiable information.

In parallel, federated learning (FL) has gained traction as a privacy-preserving paradigm that enables collaborative model training without exchanging raw patient data. However, FL in healthcare faces substantial practical challenges, including strong inter-institutional heterogeneity, severe outcome imbalance, uneven client sizes, and the lack of high-quality tabular datasets that can be openly shared for benchmarking federated optimisation strategies. Furthermore, while interpretability is critical for clinical adoption, most explainable AI techniques are not designed to operate under the 
distributed settings.

Thus, this work attempts to address several challenges in realistic synthetic data generation, robust federated optimisation under heterogeneity, and interpretability of privacy-compatible models. First, strict privacy regulations and institutional data governance severely limit the availability of sharable, real-world EHR datasets, hindering reproducible research and systematic benchmarking of FL methods. Second, while synthetic data generation has gained attention, many existing approaches for tabular clinical data fail to preserve multivariate feature dependencies or reconstruct engineered medical variables that are critical for downstream risk prediction. Third, current FL benchmarks rarely exploit synthetic data as a principled pretraining mechanism, despite its potential to stabilise optimisation under strong non-Independent and Identically Distributed (IID) client heterogeneity. Finally, 
most explainable AI techniques assume centralised access to patient-level features and are therefore incompatible with the privacy constraints of FL environments.

Motivated by these challenges, we propose a unified methodological framework that enables privacy-preserving development of clinical prediction models in realistic federated settings. The framework generates high-fidelity synthetic EHR data that capture clinically meaningful structure while avoiding patient-identifiable information, and leverages these synthetic data to warm-start FL via a synthetic pretraining phase. This initialisation conditions subsequent federated optimisation, improving robustness under non-IID client distributions. In addition, the framework integrates heterogeneity-aware
local objectives, parameter aggregation conditioned on synthetic
initialisation, post-hoc probability calibration, and privacy-compatible
model explainability to ensure that the resulting models are reliable,
interpretable, and clinically coherent across varying federation sizes.

\subsection{Contributions}

This work makes the following key contributions:

\begin{itemize}
    \item \textbf{A hybrid autoencoder–diffusion model for high-fidelity synthetic EHR generation:}The proposed generator preserves univariate, bivariate, and multivariate clinical structure, reconstructs engineered features, and demonstrates strong fidelity across distributional, correlational, and manifold-based evaluations.


\item \textbf{A synthetic-pretrained federated learning framework (SynPre-FL):}
The synthetic pretraining strategy provides a clinically meaningful global initialisation, stabilising federated optimisation, and improves performance under non-IID settings. SynPre-FL integrates local optimisation strategies, customised regularisation, synthetic-aware parameter aggregation, and post-hoc probability calibration.

\item \textbf{A privacy and utility evaluation suite for synthetic EHR assessment:} Synthetic data quality is accessed using membership inference attacks,
nearest-neighbour analysis, anomaly-based outlier detection, and
classical Train on Synthetic, Test on Real~(TSTR)/ Train on Real, Test on Synthetic (TRTS) utility tests.

    \item \textbf{A federated-safe explainability framework:}
    KernelSHAP~\cite{NIPS2017_8a20a862} is applied using privacy-preserving background selection,
    enabling global and local interpretability without exposing raw client data.
    \item \textbf{A reproducible benchmark for tabular clinical FL:}
 To our knowledge, this is among the first studies to combine
    latent diffusion–based synthetic EHR generation with FL and federated-safe explainability in a unified methodological pipeline.

\end{itemize}

\section{Background and Related Work}

    \subsection{Synthetic Data for Healthcare}

Synthetic data has emerged as a key enabler for mitigating data scarcity, privacy constraints, and regulatory barriers in AI in healthcare. Giuffre and Shung~\cite{giuffre2023harnessing} provide a broad conceptual overview of generative approaches, including GANs, VAEs, and agent-based models, highlighting their potential for privacy-preserving innovation. However, they identify critical unresolved challenges such as bias amplification, limited interpretability, inadequate auditing, and regulatory gaps under HIPAA and GDPR, motivating the need for explainability, privacy-by-design, and lifecycle governance.

Subsequent work focuses on technical limitations in realistic data synthesis. Sun et al.~\cite{SUN2023104404} address tabular healthcare data generation under class imbalance and mixed data types, proposing DP-CGANs with dependency-aware conditioning and differential privacy guarantees. Their results reveal an inherent privacy–utility trade-off and show that dependency preservation is largely confined to pairwise relationships. Li et al.~\cite{li2023generating} extend this line of work to mixed-type longitudinal EHR data, introducing EHR-M-GAN, which combines dual-VAE latent alignment with a coupled recurrent GAN architecture to better capture temporal and cross-modal dependencies, achieving improved fidelity and predictive utility at increased computational cost.

Beyond data realism, the reliability of inference on synthetic data is a growing concern. El Emam et al.~\cite{el2024evaluation} evaluate whether analyses conducted on fully synthetic datasets replicate conclusions drawn from real data, demonstrating that naive use of synthetic data can lead to misleading inferences despite low privacy risk. At a broader level, Pezoulas et al.~\cite{PEZOULAS20242892} provide a PRISMA-guided review covering tabular, time-series, imaging, omics, and multimodal healthcare data. Their taxonomy highlights the dominance of GAN- and VAE-based methods while emphasising that high distributional fidelity alone does not ensure privacy, utility, or fairness. Persistent challenges include the lack of standardised benchmarks, bias propagation, high computational cost, and regulatory ambiguity.

In overview, the literature reflects a progression from conceptual ideas to technical implementations and critical evaluation, converging on the view that synthetic data can support privacy-preserving healthcare analytics only with rigorous evaluation, replicability-aware analysis, explainability, and clear governance frameworks.

\subsection{Federated Learning for Tabular Health Data}

Ganzinger et al.~\cite{GANZINGER2023104280} introduce federated electronic data capture (fEDC), an architectural framework for collaborative clinical data collection that preserves institutional control over EHRs. Rather than enabling FL directly, fEDC achieves federation through standardised metadata exchange using CDISC ODM–encoded electronic case report forms, providing essential infrastructure for privacy-preserving, harmonised tabular data collection and reducing practical issues to future federated analytics in healthcare.

Wang et al.~\cite{WANG2022104176} propose SurvMaximin, a one-shot federated transfer learning framework for survival analysis in high-dimensional EHR data. By optimising worst-case performance across heterogeneous institutions using summary-level Cox regression statistics, the method yields transportable risk models robust to inter-site heterogeneity, missing features, and limited target-site data. Evaluation of multi-national EHR data from the 4CE COVID-19 consortium demonstrates strong generalisation, establishing a statistically grounded approach to federated clinical risk modelling.

Li et al.~\cite{LI2023104485} present FedScore, a privacy-preserving federated framework to construct interpretable and parsimonious clinical scoring systems from tabular EHR data. The approach decomposes score development into federated variable ranking, transformation, score derivation, and model selection, leveraging a communication-efficient one-shot distributed logistic regression algorithm (ODAL2). Experiments on multi-site emergency department data show performance comparable to centralised models, with improved cross-site stability and interpretability, addressing key clinical deployment requirements often overlooked in federated learning.

Thakur et al.~\cite{thakur2024knowledge} address data view heterogeneity in federated clinical settings through a knowledge abstraction and filtering–based framework. The method employs a global trainable knowledge vector with client-specific filtering modules to map heterogeneous local EHR representations into a shared latent space without manual feature alignment. Extensive evaluations of the CURIAL, eICU, and MIMIC-III datasets demonstrate consistent gains over FedAvg, Hypernetwork-based FL, LG-FedAvg, and AGAT EHR data in tabular, time-series, and graph-based EHR data.

\subsection{Explainable AI in Federated Settings}

Early work on explainability in federated environments focuses on enabling interpretation under strict privacy and data-partitioning constraints. Chen et al.~\cite{CHEN2022102474} introduce Explainable Vertical
Federated Learning (EVFL), an early explainable framework for vertical FL that provides explanations without sharing raw features. By combining a prediction-flipping objective with latent-space KL-divergence constraints and an Importance Rate metric, EVFL demonstrates that meaningful feature-level explanations can be obtained from distributed tabular data while preserving privacy.

Further works extend explainability to system-level FL deployments. Bárcena et al.~\cite{CORCUERABARCENA2023356} propose an FL as-a-Service framework for Beyond-5G/6G networks using interpretable Takagi–Sugeno–Kang fuzzy rule-based models trained via federated aggregation. Their results show near-centralised performance, transparent rule-level explanations, and reduced edge latency, highlighting the feasibility of explainable FL systems in real-world infrastructures. Explainable FL has also been explored in healthcare-specific tasks using post-hoc techniques. Briola et al.~\cite{10.1145/3655693.3660255} present a FED-XAI framework for breast cancer classification, showing that federated XGBoost and neural networks can match centralised performance while maintaining interpretability through aggregated SHAP explanations. The consistency of clinically relevant feature importance across settings suggests that post-hoc explainability remains effective under data decentralisation.

Moving beyond tabular classification, Ducange et al.~\cite{ducange2024federated} study Parkinson’s disease progression prediction, comparing interpretable-by-design federated fuzzy models with SHAP-explained federated neural networks. Their analysis reveals a trade-off between accuracy and explanation stability, with inherently interpretable models providing more consistent global explanations under non-IID data, underscoring explanation stability as a critical requirement.

Beyond model-centric explanations, Zhao et al.~\cite{zhao2024explainable} address explainability at both the model and the system-security levels in federated smart healthcare. By integrating gradient-based contribution analysis with verifiable secure aggregation, their framework supports transparent interpretation while detecting malicious or low-quality client updates, linking explainability with trust and robustness. Finally, Chaddad et al.~\cite{10085971} synthesise explainable AI, domain adaptation, and FL in medical applications, claiming that isolated adoption is insufficient for clinical deployment. The study highlights open challenges, including global explainability, handling non-IID data, and regulatory compliance, positioning explainable FL as a core component of trustworthy medical AI.

\subsection{Gaps in the Literature}

Despite advances in FL for privacy-preserving medical analytics, practical deployment remains constrained by data heterogeneity, limited local sample sizes, class imbalance, and restricted feature overlap across institutions. FL pipelines assume the existence of sufficient and well-aligned local datasets, which is often violated in real-world healthcare settings where data are sparse, non-IID, and shaped by site-specific clinical practices. In addition, explainable federated models reveal that unstable or biased local data distributions can undermine explanation consistency, model robustness, and clinical trust, even when privacy is preserved.

Synthetic data generation offers a complementary mechanism to address these limitations by enabling controlled data augmentation, distribution alignment, and representation balancing without exposing sensitive patient records. When integrated with FL, synthetic data can mitigate local data scarcity, support harmonisation between heterogeneous sites, and improve both predictive performance and explanation stability. Moreover, synthetic data can facilitate pre-training, simulation of rare conditions, and stress-testing of FL models under diverse scenarios, while remaining compatible with regulatory and privacy constraints. These considerations motivate the development of unified pipelines that combine synthetic data generation with FL and explainability, positioning synthetic-FL frameworks as a building block for scalable, trustworthy, and deployment-ready medical AI systems.


\section{Proposed Model}
\label{sec:proposed_model}
This work introduces \textbf{SynPre-FL}, a unified framework that integrates high-fidelity synthetic EHR generation with privacy-preserving FL for clinical risk prediction. SynPre-FL primarily focuses on conditioning FL through task-aware synthetic initialisation, heterogeneity-aware local objectives, and adaptive server-side optimisation.

The central idea of SynPre-FL is to decouple data availability constraints from model initialisation and optimisation. Synthetic data are first generated to capture a clinically meaningful population structure without exposing patient-identifiable information. These synthetic samples are then used to warm-start FL, placing the global model in a favourable region of the parameter space before exposure to heterogeneous client data. Subsequent federated optimisation incorporates class-balanced local losses, proximal regularisation, and adaptive aggregation to stabilise training under non-IID distributions.

SynPre-FL is designed as a modular pipeline composed of three tightly coupled stages: (i) synthetic EHR generation and validation, (ii) synthetic-pretrained federated optimisation, and (iii) post-hoc calibration and explainability. Each module can be independently analysed while contributing to the full pipeline for privacy-aware clinical prediction.
\begin{figure*}
    \centering
    \includegraphics[width=\textwidth]{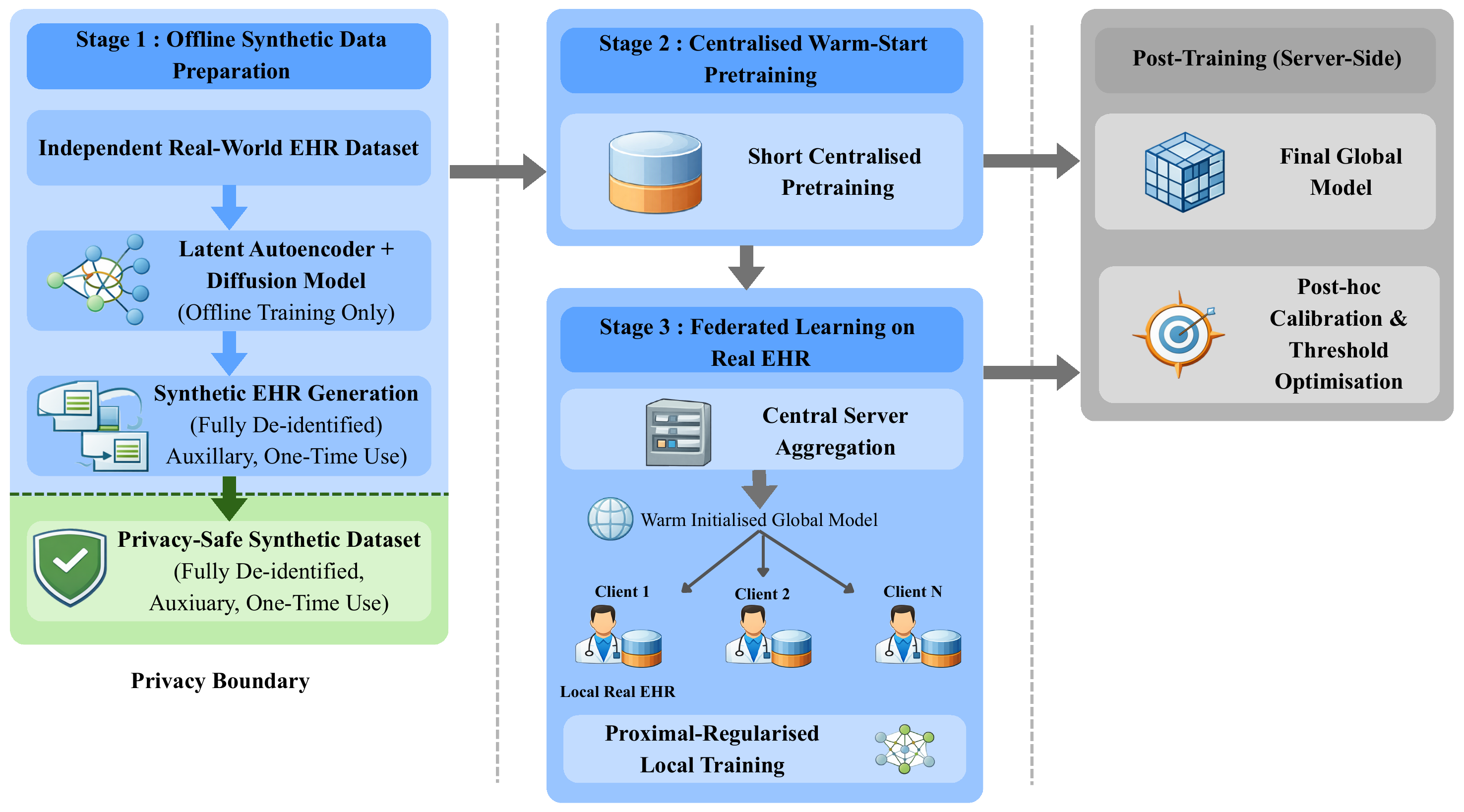}
    \caption{Stage-wise overview of the SynPre-FL framework illustrating the integration of synthetic EHR generation, synthetic pretraining, and heterogeneity-aware FL optimisation. The pipeline depicts the separation between offline synthetic data preparation and privacy-preserving FL training, followed by calibration and explainability modules for reliable risk prediction.}
    \label{fig:synpre_pipeline}
\end{figure*}

\subsection{SynPre-FL Pipeline}
\label{subsec:synpre_pipeline}

Figure~\ref{fig:synpre_pipeline} illustrates the operation of the SynPre-FL pipeline. The pipeline begins with a real-world EHR dataset that remains entirely local and is never shared between institutions. From these data, a synthetic cohort is generated using a latent autoencoder–diffusion model that preserves the multivariate clinical structure while removing patient-identifiable information. Synthetic records are completed through a task-aware label transfer mechanism, producing a fully labelled synthetic dataset compatible with the downstream prediction task.

\paragraph*{Synthetic data locality and federated constraints} Throughout the SynPre-FL pipeline, raw real-world EHR data remain strictly local to each client and are never transmitted to the server. The synthetic dataset used for pretraining is generated offline and does not correspond to identifiable patient records. Thus, it is treated as an auxiliary privacy-safe dataset available to the server before federated optimisation. Synthetic pretraining occurs only once, before federated rounds begin, and the synthetic data are not used during client-side updates. This design preserves standard FL assumptions while allowing for informed global initialisation.

In the second stage, the synthetic dataset is used to perform a short centralised pretraining phase. This synthetic pretraining step initialises the global model with clinically meaningful representations before FL training begins. Importantly, synthetic pretraining is not intended to replace learning from real data, but rather to provide a stable and privacy-safe initialisation to mitigate early-round instability under strong client heterogeneity.

In the third stage, FL is performed across multiple non-IID clients. Each client optimises a class-balanced local objective augmented with a proximal regularisation term to limit drift from the global model. Client updates are aggregated on the server using weighted averaging and adaptive Adam-style server optimization~\cite{reddi2021adaptivefederatedoptimization}. This combination improves convergence stability without altering the fundamental FL protocol.

Finally, the trained global model undergoes post-hoc probability calibration and threshold optimisation to ensure reliable clinical risk estimates. The behavior of the model is further analysed using federated-safe explainability tools that operate exclusively on the global predictor, preserving data privacy throughout the pipeline.

\subsection{Design Rationale}
\label{subsec:design_rationale}

The design of SynPre-FL is guided by three practical challenges commonly encountered in federated clinical prediction tasks: limited data shareability, strong cross-client heterogeneity, and the need for trustworthy model output.

\paragraph{Synthetic pretraining as initialisation}
In non-IID FL settings, random initialisation can lead to unstable early updates and slow convergence, particularly when client sizes and label distributions differ substantially. By pretraining the global model on task-consistent synthetic data, SynPre-FL provides an informed initialisation that captures the population-level clinical structure while remaining independent of any single client. This strategy improves the stability of the optimisation without leaking real patient information.

\paragraph{Heterogeneity-aware local optimisation} Clinical datasets are typically imbalanced and client-level label distributions vary widely under realistic partitioning. SynPre-FL addresses this through class-balanced local loss functions and proximal regularisation, reducing label imbalance issues and preventing dominance by large or skewed clients. Although these components are well-studied individually, their combination within a synthetic-pre-trained pipeline improves robustness under severe non-IID conditions.

\paragraph{Adaptive aggregation without altering the FL protocol} SynPre-FL adopts an Adam-inspired server update to adaptively scale aggregated client updates.
This formulation improves convergence smoothness and stability while preserving the standard synchronous FL structure. The proposed framework does not primarily focus on redefining aggregations; rather, it presents how adaptive aggregation is conditioned on synthetic initialisation and heterogeneity-aware local updates.

\paragraph{Clinical reliability and interpretability}
Since clinical deployments require reliable probability estimates, SynPre-FL incorporates post-hoc calibration and threshold optimisation as mandatory components of the pipeline. Explainability is performed exclusively on the global model using privacy-compatible SHAP analysis, ensuring transparency without violating federated constraints.

Together, these design choices position SynPre-FL as a practical and systematically designed framework that bridges synthetic data generation and FL for privacy-preserving clinical risk modelling.

\section{Synthetic Data Preparation Pipeline}
\label{sec:synthetic_pipeline}

This section describes the offline synthetic data generation process used to construct the auxiliary dataset for SynPre-FL.The complete synthetic generation workflow is summarised in Algorithm~\ref{alg:synthetic_generation}. The pipeline is executed prior to federated training and does not involve any interaction with federated clients. Its purpose is to generate clinically coherent, task-compatible synthetic records that can be used for model initialisation. 

\subsection{Source Dataset and Feature Scope}
The real EHR dataset~\cite{chen2022simulation} consists of $N$ patient records, from which a unified 29-feature schema was extracted comprising demographic attributes, comorbidities, respiratory symptoms, and engineered clinical risk indicators. From these records, a subset of $d_0=22$ primitive features is retained for generative modelling. Engineered features are excluded during generation and deterministically reconstructed after sampling to preserve clinical consistency.

\subsection{Preprocessing and Encoded Representation}

Let $x \in \mathbb{R}^{d_0}$ denote a real patient record. Features are partitioned into categorical variables
$\mathcal{C}$ (age group, gender, smoking category) and binary clinical indicators $\mathcal{B}$. A fixed column transformer $\phi(\cdot)$ maps the raw features into an encoded space:
\begin{equation}
z_x = \phi(x)
= \big[ \mathrm{OHE}_{\mathcal{C}}(x_{\mathcal{C}}),\; x_{\mathcal{B}} \big]
\in \mathbb{R}^{D},
\end{equation}
where $\mathrm{OHE}$ denotes one-hot encoding and $D=26$ in our implementation. The transformer $\phi$ is stored and reused for decoding synthetic samples and for all downstream evaluations.

\subsection{Latent Autoencoder}
To obtain a compact representation suitable for generative modelling, an autoencoder is trained on encoded real features.
The encoder $f_{\mathrm{enc}}:\mathbb{R}^{D}\rightarrow\mathbb{R}^{L}$ maps the encoded records to a latent space of dimension $L=32$, and the decoder $f_{\mathrm{dec}}$ reconstructs the encoded features. The autoencoder is trained by minimising the reconstruction loss:
\begin{equation}
\mathcal{L}_{\mathrm{AE}}
= \frac{1}{N} \sum_{i=1}^{N}
\left\| z_{x_i} - f_{\mathrm{dec}}(f_{\mathrm{enc}}(z_{x_i})) \right\|_2^2,
\end{equation}
using Adam with learning rate $10^{-3}$.
The trained encoder is subsequently frozen.

\subsection{Latent Diffusion Model}
A denoising diffusion model is trained directly in the latent space. Let $z_0 = f_{\mathrm{enc}}(z_x)$ denote a latent representation. A linear noise schedule $\{\beta_t\}_{t=1}^{T}$ with $T=300$ defines the forward diffusion process:
\begin{equation}
q(z_t \mid z_0)
= \mathcal{N}\!\left(
\sqrt{\bar{\alpha}_t}\, z_0,\;
(1-\bar{\alpha}_t) I
\right),
\quad
\bar{\alpha}_t = \prod_{\tau=1}^{t} (1-\beta_\tau).
\end{equation}

A neural denoiser $\varepsilon_\theta(z_t,t)$ is trained to predict injected noise by minimising:
\begin{equation}
\mathcal{L}_{\mathrm{diff}}
= \mathbb{E}_{z_0,t,\epsilon}
\left[
\left\| \epsilon - \varepsilon_\theta(z_t,t) \right\|_2^2
\right].
\end{equation}

\subsection{Synthetic Record Construction}
Synthetic latent samples are obtained by reversing the diffusion process starting from $z_T \sim \mathcal{N}(0,I)$:
\begin{equation}
z_{t-1} =
\frac{1}{\sqrt{\alpha_t}}
\left(
z_t - \beta_t \varepsilon_\theta(z_t,t)
\right).
\end{equation}
The resulting $z_0$ is decoded via $f_{\mathrm{dec}}$ and mapped back to interpretable features using inverse transformations.
Categorical attributes are recovered by argmax decoding, and binary features are thresholded. Engineered clinical features (e.g., respiratory severity, diabetes complication count, MI severity) are then recomputed deterministically using the same rules applied to real data, yielding full 29-feature synthetic records.

\subsection{Task-Aware Label Assignment}

As the generative model is unconditional, outcome labels are assigned post hoc using a label-transfer model. A gradient boosting classifier $h_{\mathrm{lab}}$ is trained on real data to estimate $p(y=1 \mid x)$.
Synthetic labels are obtained as:
\begin{equation}
\tilde{y}_j = \mathbb{I}\big(h_{\mathrm{lab}}(\tilde{x}_j) \ge 0.5\big),
\end{equation}
producing a task-consistent synthetic dataset
$\mathcal{D}_{\mathrm{syn}}$.


\section{Proposed Federated Learning Methodology}
\label{sec:fl_method}

This section details the FL training strategy used in SynPre-FL. All real patient data remain strictly local to participating clients. Synthetic data are used exclusively for global model initialisation and are not accessed during federated rounds.

\subsection{Client Partitioning}

The real dataset is partitioned into $K \in \{5,10,15\}$ non-IID client datasets with heterogeneous sizes. Each client $k$ holds local data $\mathcal{D}_k$ with $n_k$ samples. No raw data or intermediate statistics are exchanged.

\subsection{Base Prediction Model}

The global predictor is a multilayer perceptron that accepts 29 input features, followed by three hidden layers with 128, 64, and 32 neurons, respectively. The network used ReLU activations and a final sigmoid output.

\subsection{Class-Balanced Local Objective}

Each client optimises a class-balanced binary cross-entropy loss.
Let $w_+ = N_{\mathrm{neg}} / N_{\mathrm{pos}}$ be the positive-class weight.
For logit $s$ and label $y$:
\begin{equation}
\ell_{\mathrm{CB}}(s,y)
= - w_+ y \log \sigma(s)
- (1-y)\log(1-\sigma(s)).
\end{equation}

\subsection{FedProx Regularisation}
To mitigate client drift, a proximal penalty is added to the local objective:
\begin{equation}
\mathcal{L}_k(w)
= \frac{1}{n_k}\sum_{(x,y)\in\mathcal{D}_k}\ell_{\mathrm{CB}}(f_w(x),y)
+ \frac{\mu}{2}\|w - w^{(r)}\|_2^2.
\end{equation}

\subsection{Adaptive Server Optimisation}
Client updates are aggregated using weighted averaging. The server applies an Adam-style adaptive update:
\begin{align}
m^{(r)} &= \beta_1 m^{(r-1)} + (1-\beta_1) g^{(r)},\\
v^{(r)} &= \beta_2 v^{(r-1)} + (1-\beta_2) (g^{(r)})^2,\\
w^{(r+1)} &= w^{(r)} - \eta \frac{m^{(r)}}{\sqrt{v^{(r)}}+\epsilon},
\end{align}
where $g^{(r)}$ denotes the aggregated update direction.

\subsection{Synthetic Pretraining and Training Workflow}

Before federated training, the global model is pretrained on $\mathcal{D}_{\mathrm{syn}}$ for 10 epochs. FL training then proceeds for $R$ rounds.

\begin{algorithm}[t]
\caption{Synthetic EHR Generation via Autoencoder + Latent Diffusion}
\label{alg:synthetic_generation}
\begin{algorithmic}[1]
\REQUIRE Real dataset $\mathcal{D}_{\text{real}} = \{(x_i, y_i)\}_{i=1}^N$, 
number of synthetic samples $M$, latent dimension $L$, diffusion steps $T$
\STATE Fit column-transformer $\phi$ on 22 primitive real features and encode $z_{x_i} = \phi(x_i)$
\STATE Train autoencoder $(f_{\text{enc}}, f_{\text{dec}})$ on $\{z_{x_i}\}$ by minimising $\mathcal{L}_{\mathrm{AE}}$
\STATE Encode real data latents $z_{0,i} = f_{\text{enc}}(z_{x_i})$
\STATE Train diffusion model $\varepsilon_\theta$ on $\{z_{0,i}\}$ using loss $\mathcal{L}_{\mathrm{diff}}$
\STATE Train label model $h_{\text{lab}}$ on $(x_i, y_i)$ using gradient boosting
\FOR{$j = 1$ to $M$}
    \STATE Sample $z_T \sim \mathcal{N}(0, I_L)$
    \FOR{$t = T,\dots,1$}
        \STATE Compute $\varepsilon_\theta(z_t, t)$
        \STATE $z_{t-1} \gets \frac{1}{\sqrt{\alpha_t}}\!\left(z_t - \beta_t \varepsilon_\theta(z_t, t)\right)$
    \ENDFOR
    \STATE $\hat{z}_x \gets f_{\text{dec}}(z_0)$
    \STATE Decode categorical blocks of $\hat{z}_x$ via argmax; threshold binary dimensions at $0.5$ to obtain $22$ primitive features $\tilde{x}_j^{(22)}$
    \STATE Recompute 5 engineered clinical features from $\tilde{x}_j^{(22)}$ to obtain full $\tilde{x}_j \in \mathbb{R}^{29}$
    \STATE Compute $\tilde{p}_j = h_{\text{lab}}(\psi(\tilde{x}_j))$ and set $\tilde{y}_j = \mathbb{I}(\tilde{p}_j \geq 0.5)$
\ENDFOR
\STATE \textbf{return} Synthetic dataset $\mathcal{D}_{\text{syn}} = \{(\tilde{x}_j, \tilde{y}_j)\}_{j=1}^{M}$
\end{algorithmic}
\end{algorithm}

\begin{algorithm}[t]
\caption{SynPre-FL: Synthetic Pretraining + Federated Optimisation}
\label{alg:synpre_fl}
\begin{algorithmic}[1]
\REQUIRE Real client datasets $\{\mathcal{D}_k\}_{k=1}^{K}$, synthetic dataset $\mathcal{D}_{\text{syn}}$, number of rounds $R$, local epochs $E$, server learning rate $\eta$
\STATE Initialise global model weights $w^{(0)}$ randomly
\STATE \textbf{Synthetic pretraining:}
\STATE \quad Train $w^{(0)}$ on $\mathcal{D}_{\text{syn}}$ for 10 epochs with class-balanced BCE to obtain $w^{(0)}_{\text{pre}}$
\STATE Set $w^{(0)} \leftarrow w^{(0)}_{\text{pre}}$, initialise $m^{(0)}=0$, $v^{(0)}=0$
\FOR{round $r = 0$ to $R-1$}
    \FOR{each client $k = 1,\dots,K$ \textbf{in parallel}}
        \STATE Receive current global model $w^{(r)}$
        \STATE Set local weights $w_k \leftarrow w^{(r)}$
        \FOR{$E$ local epochs}
            \FOR{mini-batch $(u,y) \subset \mathcal{D}_k$}
                \STATE Compute class-balanced FedProx loss $\mathcal{L}_k(w_k; w^{(r)})$
                \STATE Update $w_k$ with Adam on $\nabla_{w_k} \mathcal{L}_k$
            \ENDFOR
        \ENDFOR
        \STATE Send updated weights $w_k^{(r)} \leftarrow w_k$ to server
    \ENDFOR
    \STATE Compute weighted average $\bar{w}^{(r)} = \sum_{k} \tfrac{n_k}{n} w_k^{(r)}$
    \STATE Form pseudo-gradient $g^{(r)} = w^{(r)} - \bar{w}^{(r)}$
    \STATE Update Adam moments $m^{(r)}, v^{(r)}$ and compute $w^{(r+1)}$ as in FedAdam
\ENDFOR
\STATE \textbf{return} Final global model $w^{(R)}$
\end{algorithmic}
\end{algorithm}

\begin{algorithm}[t]
\caption{Calibration and Threshold Optimisation}
\label{alg:calibration}
\begin{algorithmic}[1]
\REQUIRE Trained global model $f_{w^{(R)}}$, validation set $\mathcal{V}$
\STATE Compute raw probabilities $p_i = f_{w^{(R)}}(u_i)$ for all $(u_i, y_i) \in \mathcal{V}$
\STATE Fit logistic recalibration model $(a,b)$ on $(p_i, y_i)$
\STATE Obtain calibrated scores $\tilde{p}_i = \sigma(a \log(p_i/(1-p_i)) + b)$
\STATE Define a grid of thresholds $\mathcal{T} \subset [0,1]$
\FOR{each $\tau \in \mathcal{T}$}
    \STATE $\hat{y}_i^{(\tau)} \gets \mathbb{I}(\tilde{p}_i \ge \tau)$
    \STATE Compute $\mathrm{F1}(\tau)$ on $\mathcal{V}$
\ENDFOR
\STATE $\hat{\tau} \gets \arg\max_{\tau \in \mathcal{T}} \mathrm{F1}(\tau)$
\STATE \textbf{return} Calibrated decision rule $(f_{w^{(R)}}, \hat{\tau})$
\end{algorithmic}
\end{algorithm}


\section{Probability Calibration}
\label{sec:calibration}

Existing studies have shown that neural network classifiers produce poorly calibrated probability
estimates, particularly with class imbalance, heterogeneous training
distributions, and non-convex optimisation dynamics. In the FL context, these effects become more evident where client-specific data biases and non-IID aggregation
can distort the global logit scale. Since the SynPre-FL model is intended for
risk-sensitive clinical decision support, post-hoc probability calibration is
required to ensure that output scores correspond to reliable empirical event
likelihoods. The calibration and threshold optimisation procedure used in this study is summarised in Algorithm~\ref{alg:calibration}.

\subsection{Need for Calibration in Health Models}

Medical classification tasks often require threshold-based decision-making
(e.g., identifying high-risk patients), where the reliability of the predicted probabilities directly influences clinical utility and fairness. Despite strong discriminative performance, federated neural models frequently exhibit systematic miscalibration due to:
\begin{itemize}
    \item \textbf{Label imbalance}, which biases sigmoid outputs toward the majority class.
    \item \textbf{Client-level heterogeneity}, which leads to inconsistent logit scaling.
    \item \textbf{Non-IID aggregation effects} that perturb global decision boundaries.
    \item \textbf{Distributional shifts induced by synthetic initialisation}.
\end{itemize}
These issues jointly motivate a principled calibration mechanism applied to the final global model after completion of the full SynPre-FL training pipeline.

\subsection{Logistic Recalibration}
We adopt a Platt-style logistic recalibration scheme~\cite{van2019calibration,guo2017calibration}, adding a lightweight logistic regression over the raw logits of the final global model using a held-out real-data validation set. Given the uncalibrated logit output $f(x)$ from the final global model, the calibrated probability is defined as:
\begin{equation}
    \tilde{p}(x) = \sigma(a f(x) + b),
\end{equation}
where $\sigma(\cdot)$ denotes the sigmoid function, and the parameters $(a,b)$ are estimated by minimising the negative log-likelihood over the validation split. Compared to temperature scaling, logistic recalibration introduces an additive bias term and is better suited for tabular models whose logit distributions may not be symmetric or unimodal.

Calibration was performed strictly on the validation set to avoid leakage of test information, and the resulting calibrated model was used for all downstream threshold selection and evaluation.

\subsection{Threshold Optimisation for F1}

After obtaining calibrated probability scores, we perform grid search over candidate thresholds $T \in [0.1,0.9]$ to identify the operating point that maximises the F1 score on the validation set. Extreme thresholds near 0 and 1 were excluded to avoid unstable metric estimates under strong class imbalance. The optimal threshold is defined as
\begin{equation}
    T^{*} = \arg\max_{T \in [0.1,0.9]} \mathrm{F1}(T),
\end{equation}
where $\mathrm{F1}(T)$ is computed by thresholding the calibrated probabilities
$\tilde{p}(x)$ at level $T$.

The selected threshold $T^{*}$ is fixed after validation and applied during evaluation on the held-out test data, producing the calibrated metrics reported in Sec.~\ref{sec:results}.



\section{Explainability Framework}
\label{sec:explainability}

The \textbf{SynPre-FL} framework incorporates a privacy-preserving explainability module based on SHAP (Shapley additive explanations), enabling both global and local interpretability without exposing patient-level records. This section details the explainability constraints, the SHAP attribution pipeline, and the methodology for aggregating explanations across distributed clients.

\subsection{Federated Explainability Constraints}

In FL, as the raw patient features and labels remain localised, 
explainability methods must operate under the following constraints:

\begin{itemize}
    \item \textbf{No access to raw client data:} Attribution methods cannot rely on
          centralised pooling of individual patient records.
    \item \textbf{Model-centric explainability:} Explanations must be derived solely
          from the global model, model outputs, and optional privacy-safe surrogate
          data.
    \item \textbf{Client invariance:} Explanations should remain stable across varying client counts ($5, 10, 15$), reflecting consistent global reasoning despite data fragmentation.
\end{itemize}

These constraints motivate a federated-safe explainability pipeline that enables transparent model auditing while maintaining strict privacy guarantees.

\subsection{KernelSHAP-Based Attribution}

KernelSHAP was selected for explainability due to its model-agnostic nature, robustness to heterogeneous feature interactions, and compatibility with federated privacy requirements. SHAP values were computed on the raw probability outputs of the global model prior to post-hoc calibration. Calibration was applied only for decision threshold optimisation and evaluation, while explainability analysis focused on the underlying predictive behaviour of the trained model.

\subsubsection*{\textbf{Background Dataset Selection}}

SHAP requires a reference distribution to estimate conditional expectations. To obtain a privacy-safe and computationally efficient background set, a fixed subset of 200 representative samples from a global held-out real dataset was used as the background distribution. This ensures that key regions of the feature space are covered while preventing data leakage from individual clients. 

\subsubsection*{\textbf{Global SHAP Aggregation}}

A set of 300 evaluation samples was used to compute SHAP values for the calibrated global model. For each sample, KernelSHAP estimates a vector of feature contributions
$\phi_j$ such that:
\[
f(x) = f(x_{\text{baseline}}) + \sum_{j=1}^{d} \phi_j,
\]
where $d$ is the dimensionality of the input feature vector after deterministic preprocessing.

Global feature importance is then computed as:
\[
\mathrm{Imp}(j) = \frac{1}{N} \sum_{i=1}^{N} |\phi_{i,j}|,
\]
where $N=300$ in our evaluations.

This aggregation captures the dominant predictors learned collectively from all federated clients, enabling direct comparison of feature relevance across different federation sizes.

\subsubsection*{\textbf{Top-Features Ranking}}

Features were ranked using their mean absolute SHAP contributions. Across all client configurations, a consistent set of clinically meaningful predictors emerged, including:

\begin{itemize}
    \item \textbf{Gender}\vspace{-0.4em}
    \item \textbf{Age category}\vspace{-0.4em}
    \item \textbf{Ischemic heart disease (IHD)}\vspace{-0.4em}
    \item \textbf{Smoking severity}\vspace{-0.4em}
    \item \textbf{Diabetic conditions}\vspace{-0.4em}
    \item \textbf{Respiratory distress, cough, and dyspnea}\vspace{-0.4em}
    \item \textbf{Hypertension}
\end{itemize}

The consistency of these rankings across the 5-, 10-, and 15-client SynPre-FL models demonstrates that the federated training process preserves underlying clinical relationships captured by the model, even under severe data fragmentation and heterogeneity. 

\section{Results}\label{sec:results}
This section presents a comprehensive evaluation of the proposed synthetic EHR generation pipeline and its downstream application to federated learning. We report three main categories of analyses: (i) fidelity, (ii) privacy, and (iii) utility. We further demonstrate the impact of synthetic pretraining under multi-client FL using 5, 10, and 15 heterogeneous clients, followed by model calibration and explainability analyses.  




\subsection{Synthetic Data Generation Performance}

\subsubsection*{\textbf{Autoencoder Training}}
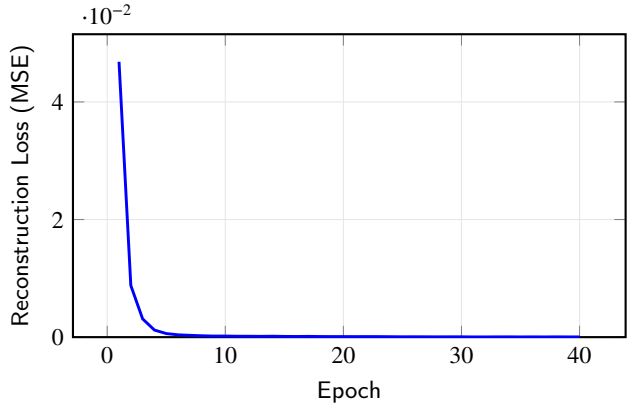
\begin{figure}
\centering
\begin{tikzpicture}
\begin{axis}[
    width=3.5in,
    height=2.2in,
    xlabel={Epoch},
    ylabel={Reconstruction Loss (MSE)},
    ymin=0,
    grid=both,
    grid style={gray!20},
    thick,
]
\addplot[
    blue,
    line width=1.1pt,
]
coordinates {
(1,0.046829)
(2,0.008784)
(3,0.003106)
(4,0.001206)
(5,0.000610)
(6,0.000395)
(7,0.000307)
(8,0.000235)
(9,0.000182)
(10,0.000184)
(11,0.000155)
(12,0.000152)
(13,0.000134)
(14,0.000155)
(15,0.000125)
(16,0.000111)
(17,0.000133)
(18,0.000104)
(19,0.000095)
(20,0.000094)
(21,0.000089)
(22,0.000095)
(23,0.000089)
(24,0.000073)
(25,0.000058)
(26,0.000064)
(27,0.000057)
(28,0.000050)
(29,0.000056)
(30,0.000055)
(31,0.000035)
(32,0.000036)
(33,0.000046)
(34,0.000049)
(35,0.000036)
(36,0.000046)
(37,0.000036)
(38,0.000048)
(39,0.000039)
(40,0.000035)
};
\end{axis}
\end{tikzpicture}
\caption{Training loss of the autoencoder used for latent representation learning within the synthetic EHR generation pipeline. The reduction in reconstruction loss across 40 epochs indicates stable convergence}
\label{fig:ae_loss}
\end{figure}

The autoencoder was trained for 40 epochs. As presented in Figure~\ref{fig:ae_loss}, reconstruction loss showed rapid convergence, reaching a final MSE of $3.5 \times 10^{-5}$.  

\subsubsection*{\textbf{Diffusion Model Training}}
\begin{figure}
\centering
\begin{tikzpicture}
\begin{axis}[
    width=3.5in,
    height=2.2in,
    xlabel={Epoch},
    ylabel={Diffusion Training Loss},
    ymin=0.45,
    grid=both,
    grid style={gray!20},
    thick,
]
\addplot[
    red,
    line width=1.1pt,
]
coordinates {
(1,0.917649)
(2,0.765708)
(3,0.721146)
(4,0.703855)
(5,0.691797)
(6,0.676610)
(7,0.662340)
(8,0.655362)
(9,0.643925)
(10,0.635290)
(11,0.631429)
(12,0.620222)
(13,0.614483)
(14,0.606777)
(15,0.598825)
(16,0.595430)
(17,0.585888)
(18,0.581690)
(19,0.572414)
(20,0.573997)
(21,0.570852)
(22,0.562428)
(23,0.560993)
(24,0.553823)
(25,0.545798)
(26,0.546156)
(27,0.539234)
(28,0.537537)
(29,0.534941)
(30,0.530195)
(31,0.526417)
(32,0.524495)
(33,0.517332)
(34,0.514226)
(35,0.512188)
(36,0.507329)
(37,0.502832)
(38,0.502315)
(39,0.498824)
(40,0.494352)
};
\end{axis}
\end{tikzpicture}
\caption{Training loss of the latent diffusion model over 40 epochs. The gradual decrease demonstrates successful learning of the latent feature distribution and stable optimisation behaviour during synthetic data generation.}
\label{fig:diff_loss}
\end{figure}
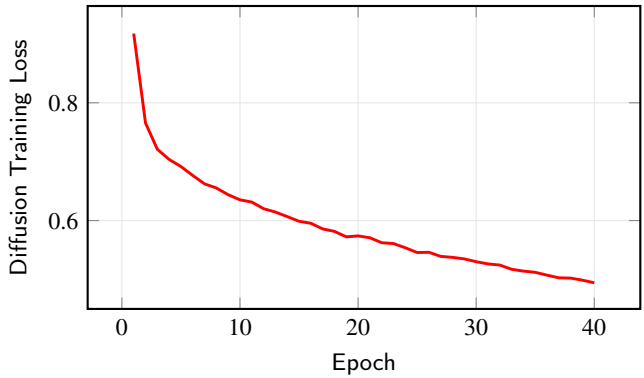

The latent diffusion model similarly converged as expected, reducing training loss from $0.92$ to $0.49$ (Figure~\ref{fig:diff_loss}).

\subsection{Fidelity Evaluation}
We evaluated fidelity using three complementary analysis methods: (i) univariate distribution similarity, (ii) bivariate correlation structure, and (iii) multivariate manifold similarity. These analyses quantify how closely the synthetic EHR data reproduces the statistical properties of the real dataset.

\subsubsection*{\textbf{Univariate Fidelity}}
For univariate analysis, the categorical Jensen--Shannon Divergence (JSD)~\cite{dorent2025connectingjensenshannonkullbackleiblerdivergences} value was computed between the real and synthetic marginal distributions for all categorical and binary features. As shown in Table~\ref{tab:jsd}, all JSD values remain below 0.20, indicating strong alignment between real and synthetic univariate distributions, with the lowest divergence observed in demographic variables (e.g., gender, age) and slightly higher divergence in rare disease indicators.

\begin{table}
\centering
\caption{JSD values for categorical/binary features. Lower values indicate closer alignment between real and synthetic marginal distributions.}
\label{tab:jsd}
\begin{tabular}{l c}
\hline
\textbf{Feature} & \textbf{JSD} \\
\hline
age 30--50 & 0.0986 \\
age 50--70 & 0.1195 \\
age $>$70 & 0.0433 \\
gender (f/m) & 0.0095 \\
smoking former & 0.0277 \\
smoking never & 0.0271 \\
COPD & 0.0083 \\
emphysema & 0.0708 \\
hypertension & 0.0808 \\
respiratory cough & 0.1074 \\
dm retinopathy (proliferative) & 0.1836 \\
dm macular edema & 0.1358 \\
IHD & 0.0060 \\
\hline
\end{tabular}
\end{table}
\subsubsection*{\textbf{Bivariate Fidelity}}
\begin{figure*}
    \centering
    \includegraphics[width=\textwidth]{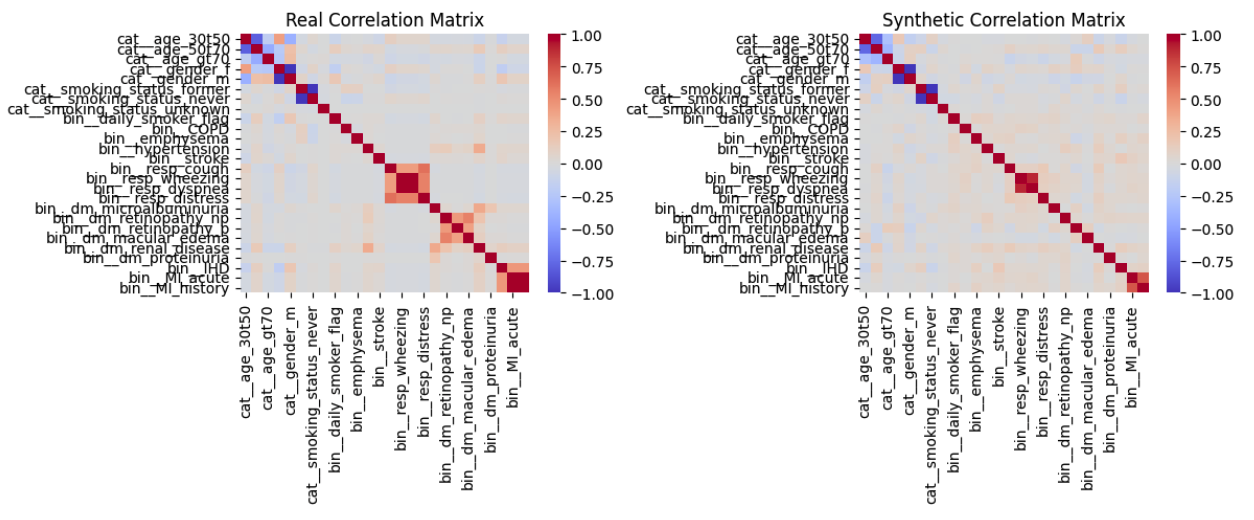}
    \caption{Comparison of real vs. synthetic correlation matrices. The synthetic data preserves major correlation structures, including clusters of respiratory symptoms and diabetic complications.}
    \label{fig:corr}
\end{figure*}

Bivariate fidelity was analysed by comparing the correlation structure of real and synthetic datasets. Figure~\ref{fig:corr} shows heatmaps of Pearson correlation matrices. The global difference between these matrices, measured using the Frobenius norm, was:
\begin{equation*}
    || C_{\text{real}} - C_{\text{syn}} ||_F = 2.5488
\end{equation*}

A lower value indicates better preservation of pairwise dependencies. Visual inspection confirms that major correlation blocks (e.g., respiratory and diabetic complication clusters) are well preserved.

\subsubsection*{\textbf{Multivariate Fidelity}}
\begin{figure}
    \centering
    \includegraphics[width=\columnwidth]{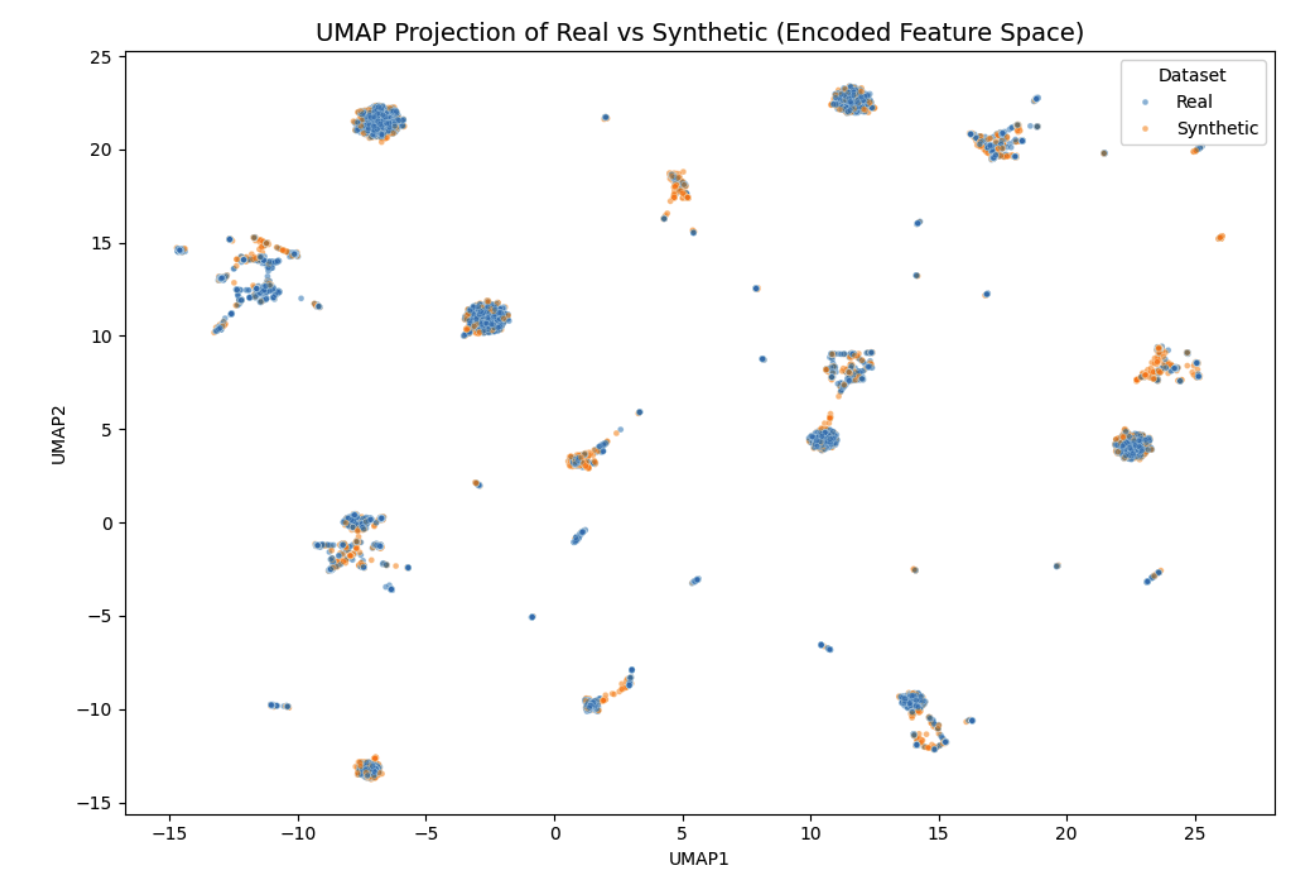}
    \caption{UMAP projection of real vs. synthetic samples in latent feature space. The strong overlap across clusters indicates preservation of global multivariate structure.}
    \label{fig:umap}
\end{figure}

To evaluate high-dimensional manifold similarity, we trained a propensity-score classifier to distinguish real from synthetic samples. The classifier achieved an AUROC of 0.7043 , indicating moderate distinguishability but substantial overlap of joint distributions. Figure~\ref{fig:umap} shows a UMAP projection of the encoded feature space, where real and synthetic samples form overlapping clusters, demonstrating that the diffusion model captures the global structure of patient phenotypes.

\subsection{Privacy Evaluation}
We evaluated privacy using two complementary approaches: (i) Membership Inference Attack (MIA) and (ii) nearest-neighbour (NN)  distance analysis in the encoded latent space. 

\paragraph*{\textbf{Membership Inference Attack}}
A classifier was trained to distinguish whether a given sample belonged to the real training set. The resulting AUROC was 0.5064, very close to random guessing (0.50), indicating no detectable membership leakage.

\paragraph*{\textbf{Nearest-Neighbor Distance Analysis}}
We computed the mean NN distance between samples in the encoded latent space:
\begin{itemize}
    \item Real with Real mean distance: 0.0718 \vspace{-0.4em}
    \item Synthetic with Real mean distance: 1.3816
\end{itemize}
The ratio of synthetic-to-real NN distance was 19.23, indicating that synthetic points lie significantly farther from real samples than real samples lie from each other. This suggests strong protection against reconstruction or linkage attacks.


%
\subsection{Utility Evaluation}

To assess the usefulness of the generated synthetic EHR data for further predictive modelling, a comprehensive utility analysis was performed, consisting of (i) classical TSTR/TRTS evaluation, and (ii) ML model-based utility. All experiments were conducted using real and synthetic datasets encoded into a 26-dimensional latent feature space.

\subsubsection*{\textbf{Classical Utility Evaluation}}

The synthetic dataset was first evaluated using the standard Train-on-Synthetic, Test-on-Real and Train-on-Real, Test-on-Synthetic protocols. Results are shown in Table~\ref{tab:utility_basic}. Synthetic data achieved strong generalisation to real test cases, while TRTS performance achieved high discrimination due to the synthetic data being less noisy.

\begin{table}
\centering
\caption{Classical utility evaluation using baseline model.}
\label{tab:utility_basic}
\begin{tabular}{l c c c}
\hline
\textbf{Setting} & \textbf{AUC} & \textbf{ACC} & \textbf{F1} \\
\hline
Real with Real & 0.8858 & 0.8432 & 0.6696 \\
TSTR & 0.8354 & 0.8557 & 0.6011 \\
TRTS & 0.9968 & 0.9316 & 0.8346 \\
\hline
\end{tabular}
\end{table}

\subsubsection*{\textbf{Model-Based Utility Evaluation}}

To further assess robustness, we evaluated synthetic data utility using categories of ML: XGBoost, a shallow neural network with focal loss and batch normalisation, and a deep neural network. Each model was trained under the same three settings as the previous test. For the deep network, we additionally evaluated a mixed data distribution validated against real data. The comparative performance across model families and evaluation settings is summarised in Table~\ref{tab:xgb}.

The fully connected deep neural network is designed for tabular clinical data, consisting of four sequential linear layers of widths 128, 64, 32, and 1. Each hidden layer uses ReLU activation to introduce non-linearity and allow the model to capture complex interactions among heterogeneous EHR features. The final layer outputs a single logit representing the predicted probability of the clinical outcome, and training is performed using a class-balanced binary cross-entropy loss to counter the substantial positive–negative label imbalance present in the real dataset. This architecture is intentionally lightweight to facilitate efficient deployment across resource-constrained clients while remaining expressive enough to model nonlinear relationships common in high-dimensional clinical phenotypes. The same model will also be used for further experimentation in the SynPre-FL pipeline.

\begin{table}
\centering
\caption{Utility Evaluation Results }
\label{tab:xgb}
\begin{tabular}{l c c c}
\hline
\textbf{Setting} & \textbf{AUC} & \textbf{ACC} & \textbf{F1} \\
\hline
\multicolumn{4}{c}{\textbf{\textit{XGBoost}} } \\
\hline
Real with Real & 0.8806 & 0.7696 & 0.6336 \\
TSTR & 0.8611 & 0.8355 & 0.6616 \\
TRTS & 0.9833 & 0.7112 & 0.5510 \\
\hline
\multicolumn{4}{c}{\textbf{\textit{Shallow Neural Network }}} \\
\hline
Real with Real & 0.8598 & 0.2441 & 0.3924 \\
TSTR & 0.8268 & 0.2441 & 0.3924 \\
TRTS & 0.9136 & 0.1790 & 0.3036 \\

\hline
\multicolumn{4}{c}{\textbf{\textit{Deep neural network}} } \\
\hline
Real with Real & 0.8646 & 0.8382 & 0.6657 \\
TRTS & 0.9143 & 0.8852 & 0.6877 \\
Mixed with Real & 0.8650 & 0.8429 & 0.6686 \\
\hline

\end{tabular}
\end{table}

Rather than memorisation or privacy leakage, the consistently high TRTS performance across all model families reflects the reduced label noise and smoother decision boundaries present in the synthetic dataset, which is generated to preserve clinically relevant structure while maintaining necessary variability as in real-world records. This interpretation is further supported by the privacy evaluations in Section~\ref{sec:results}, where membership inference attacks achieve near-random performance and NN distance analysis confirms strong separation between real and synthetic samples. Together, these results indicate that high TRTS scores arise from distributional regularisation rather than unintended information leakage.

\subsubsection*{\textbf{Summary of Utility Findings}}
Across all models, synthetic data exhibited strong predictive utility:

\begin{itemize}
    \item TSTR performance closely approached Real with Real performance, with utility retention between 0.94 and 0.98 across models.
    \item TRTS performance was consistently high, reflecting that synthetic data preserves discriminative structure while being less noisy.
    \item Deep models confirmed that synthetic and mixed data do not degrade performance when evaluated on real test sets.
\end{itemize}

These results collectively demonstrate that the generated synthetic EHR data is highly usable for downstream machine learning tasks and preserves clinically relevant predictive structure.

\subsection{Outlier Detection Analysis}
Outlier detection forms an important component of synthetic data validation, as high-quality synthetic datasets should reproduce not only the central distribution of the real population but also its rare and extreme cases.  Thus, the synthetic dataset is evaluated against the real dataset using two commonly applied unsupervised anomaly detectors: Isolation Forest (IF) and Local Outlier Factor (LOF). For each method, we report the outlier proportion in real data ($p_{\text{real}}$), the proportion in synthetic data ($p_{\text{syn}}$), and their absolute difference (Outlier Proportion Difference- OPD).

\paragraph*{\textbf{Isolation Forest}}
Real and synthetic outlier rates were $0.0100$ and $0.0236$ respectively, yielding an OPD of $0.0136$.  This low OPD indicates that the synthetic dataset reproduces the real-data anomaly structure with high fidelity.  Even though the generative model injects stochastic variability to preserve privacy, resulting in deviations, the deviation remains within the expected range. 

\paragraph*{\textbf{Local Outlier Factor}}
LOF yielded real and synthetic outlier rates of $0.0099$ and $0.0344$, corresponding to an OPD of $0.0245$.  
A slightly higher deviation is expected since LOF remains sensitive to local density fluctuations, which can vary
modestly across synthetic samples.  However, this pattern confirms that synthetic data does not contain more anomalies than the real dataset.

Thus, as shown in Table~\ref{tab:out}, both methods produce OPD values below $0.03$, indicating that the synthetic data does not introduce extreme outlier patterns and carefully reproduces the rare case structure observed in real data.

\begin{table}
\centering
\caption{Outlier Proportion Difference (OPD) between real and synthetic data.}
\begin{tabular}{lccc}
\hline
\label{tab:out}

\textbf{Method} & $p_{\text{real}}$ & $p_{\text{syn}}$ & \textbf{OPD} \\
\hline

Isolation Forest (IF) & 0.0100 & 0.0236 & \textbf{0.0136} \\
Local Outlier Factor (LOF) & 0.0099 & 0.0344 & \textbf{0.0245} \\
\hline

\end{tabular}
\end{table}

Combining the findings of the distributional comparisons, utility evaluations (TSTR/TRTS), privacy analyses, and the outlier results, the experiments validate the usability of the synthetic data generation pipeline.  The low OPD values obtained from IF and LOF ($<0.03$) align closely with the distributional fidelity and classification utility measurements, confirming that the synthetic data preserve both central and peripheral structure of the real population. Overall, the synthetic data exhibit strong alignment with real-world structure across all reliable DOTS axes, demonstrating that the generative pipeline produces usable, coherent, and privacy-conscious synthetic clinical data.


\section{Federated Learning Analysis}
This section evaluates the proposed \textbf{SynPre-FL} framework, which integrates (i) synthetic pretraining, (ii) heterogeneous multi-client federated learning, (iii) class-balanced local optimisation, (iv) FedProx regularisation, (v) Adam-based adaptive aggregation, and (vi) post-hoc probability calibration and threshold optimisation. Results are reported for \textbf{5}, \textbf{10}, and \textbf{15} client configurations, reflecting increasing levels of heterogeneity and data fragmentation.

\subsection{Dataset Description}
\label{subsec:dataset}
All experiments in this study are conducted using a publicly available EHR dataset generated with the \textbf{Synthea} patient simulator for lung cancer risk prediction \cite{chen2022simulation}. The data comprise populations of up to 150{,}000 patients, from which subsets containing lung cancer cases and matched controls were constructed.  Continuous variables are discretised into categorical values, yielding a fully tabular representation suitable for standard classifiers. The dataset was subjected to an initial screening and feature engineering to develop the 29-feature schema consisting of demographic attributes, smoking-related variables, comorbidities, respiratory symptoms, and derived clinical risk indicators. The prediction task is a binary classification, where the label indicates lung cancer presence.

We restrict our study to a single dataset because publicly shareable EHR datasets with such high level of clinical structure, feature completeness, and compatibility with FL remain extremely limited. Since the primary goal of this work is to evaluate synthetic pretraining and federated optimisation strategies rather than population-level generalizability, focusing on a well-established synthetic benchmark is appropriate and sufficient for methodological analysis.

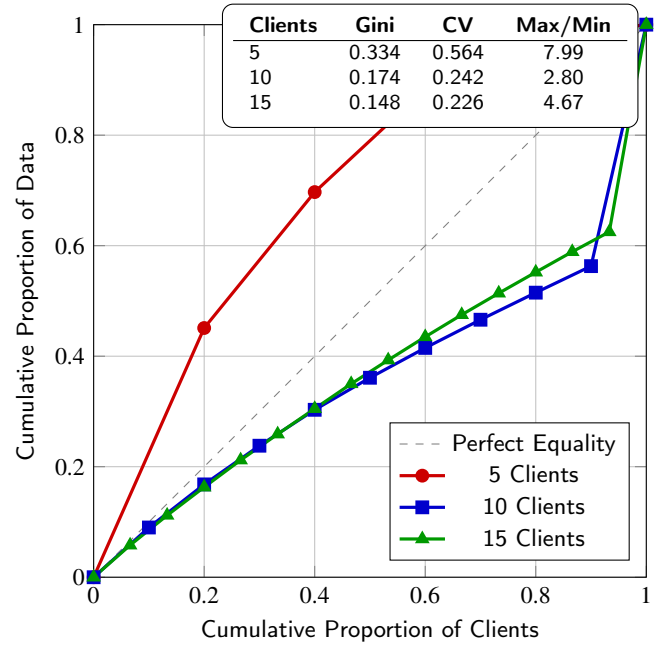
\begin{figure}
\centering

\begin{tikzpicture}[>=stealth]

\begin{axis}[
    width=3.5in,
    height=3.5in,
    xlabel={Cumulative Proportion of Clients},
    ylabel={Cumulative Proportion of Data},
    xmin=0, xmax=1,
    ymin=0, ymax=1,
    grid=both,
    grid style={line width=.1pt, draw=gray!20},
    major grid style={line width=.2pt, draw=gray!50},
    legend style={font=\small, at={(0.97,0.03)},anchor=south east},
    tick label style={font=\small},
]

\addplot[dashed, thin, gray] coordinates {(0,0) (1,1)};
\addlegendentry{Perfect Equality}

\addplot[very thick, red!80!black, mark=*] 
coordinates {
(0,0)
(0.2,0.451)
(0.4,0.697)
(0.6,0.883)
(0.8,0.973)
(1.0,1.000)
};
\addlegendentry{5 Clients}

\addplot[very thick, blue!80!black, mark=square*]
coordinates {
(0,0)
(0.1,0.090)
(0.2,0.168)
(0.3,0.238)
(0.4,0.303)
(0.5,0.361)
(0.6,0.415)
(0.7,0.466)
(0.8,0.515)
(0.9,0.563)
(1.0,1.000)
};
\addlegendentry{10 Clients}

\addplot[very thick, green!60!black, mark=triangle*]
coordinates {
(0,0)
(0.066,0.058)
(0.133,0.112)
(0.200,0.163)
(0.266,0.212)
(0.333,0.259)
(0.400,0.305)
(0.466,0.350)
(0.533,0.393)
(0.600,0.435)
(0.666,0.475)
(0.733,0.514)
(0.800,0.552)
(0.866,0.589)
(0.933,0.625)
(1.000,1.000)
};
\addlegendentry{15 Clients}

\end{axis}

\node[anchor=north east, xshift=-5pt, yshift=5pt] at (current bounding box.north east) {
\begin{tikzpicture}
\node[
    draw,
    fill=white,
    rounded corners,
    inner sep=4pt,
    font=\footnotesize
]{
\begin{tabular}{lccc}
\textbf{Clients} & \textbf{Gini} & \textbf{CV} & \textbf{Max/Min} \\
\hline
5  & 0.334 & 0.564 & 7.99 \\
10 & 0.174 & 0.242 & 2.80 \\
15 & 0.148 & 0.226 & 4.67 \\
\end{tabular}
};
\end{tikzpicture}
};

\end{tikzpicture}

\caption{Lorenz curves and heterogeneity statistics for the 5-, 10-, and 15-client SynPre-FL configurations.  
Smaller Gini coefficients in 10- and 15-client settings indicate more evenly divided total data volume, but the longer left-tail in the 15-client curve reflects greater fragmentation—capturing the rise in statistical heterogeneity despite improved balance.}
\label{fig:lorenz_heterogeneity}
\end{figure}

\paragraph*{\textbf{Client Heterogeneity}}
Real EHR datasets were partitioned into $K=\{5,10,15\}$ non-IID shards to simulate imbalanced client sizes proportional to realistic cross-institution heterogeneity. The client configurations is visualised using Lorenz curves and inequality statistics in Fig.~\ref{fig:lorenz_heterogeneity}. For instance, data distribution for the 15-client setting includes 3193, 2737, 2281, 2052, 1824, 1596, 1368, 1368, 1140, 1140, 912, 912, 912, 684, 692 samples per client. Each client retained its private subset throughout training, and no raw data were shared at any stage. 

\paragraph*{\textbf{Federated Optimisation}}
At the beginning of each communication round, the server transmitted the current global model to all clients. Then, every client initialised from the current global model and performed a local update on its private dataset using mini-batch gradient descent. We ran a total of 10 global communication rounds for all FL configurations. After federated training, the final global model was optionally fine-tuned on the full real training set to assess performance recovery. To analyse the effect of aggregation protocol on training convergence, different aggregation protocols were employed:
\begin{itemize}
    \item \textbf{Weighted FedAvg} using updated variant of the classical local SGD aggregation.
    \item \textbf{Hybrid Aggregation :} The aggregation approach integrated features of FedProx and FedAdam, combining proximal regularisation with Adam-based optimisation for improved stability under heterogeneity.
    \item \textbf{SynPre-FL}: The proposed pipeline combines a synthetic data-based pre-training mechanism with a customised FL aggregation framework.
\end{itemize}

\paragraph*{\textbf{Synthetic Pretraining (SynPre-FL)}}
Before federated training, the global model was pretrained for \textbf{10 epochs} on a synthetically generated dataset (5{,}000 samples) obtained from the proposed diffusion–autoencoder pipeline. This pretraining was designed to improve initialisation and convergence stability. The full training workflow is discussed in Algorithm~\ref{alg:synpre_fl}.

\subsection{Synthetic Pretraining Performance}
Table~\ref{tab:synpretrain} summarises model performance during the synthetic pretraining stage.  
Across epochs, the model exhibits a clear and monotonic improvement in AUROC, accuracy, and F1, indicating that the synthetic dataset carries sufficient signal to produce a meaningful initialisation prior to federated optimisation.

\begin{table}
\centering
\caption{Synthetic pretraining performance (10 epochs).}
\label{tab:synpretrain}
\begin{tabular}{c c c c c c}
\hline
Epoch & AUC & ACC & Precision & Recall & F1 \\
\hline
1 & 0.7801 & 0.7024 & 0.7620 & 0.6881 & 0.7231 \\
5 & 0.8469 & 0.7488 & 0.8015 & 0.7102 & 0.7534 \\
10 & 0.8844 & 0.7794 & 0.8324 & 0.7337 & 0.7796 \\
\hline
\end{tabular}
\end{table}

The steady improvement confirms that synthetic pretraining effectively brings the model into a favourable optimisation region before federated training begins.  Importantly, we intentionally restricted pretraining to only 10 epochs. This design choice serves two purposes:

\begin{itemize}
    \item \textbf{Preventing overfitting to synthetic artefacts :} Excessive optimisation on purely synthetic data risks embedding synthetic-specific patterns into model weights, which could negatively impact real-client updates. A short warm-up reduces this risk while still providing informative initialisation.

    \item \textbf{Maintaining computational efficiency:} The pretraining phase is executed once, before federated rounds start. Running it substantially longer would increase total computation without proportionate gain in downstream performance, especially since the purpose of pretraining is not to maximise synthetic accuracy but to stabilise the start of FL.
\end{itemize}

Thus, the synthetic pretraining phase is used as a lightweight initialisation mechanism rather than a full-fledged training stage, ensuring that the subsequent federated optimisation remains the primary driving factor of real-world performance.

\subsection{Client-Level Performance and Heterogeneity}

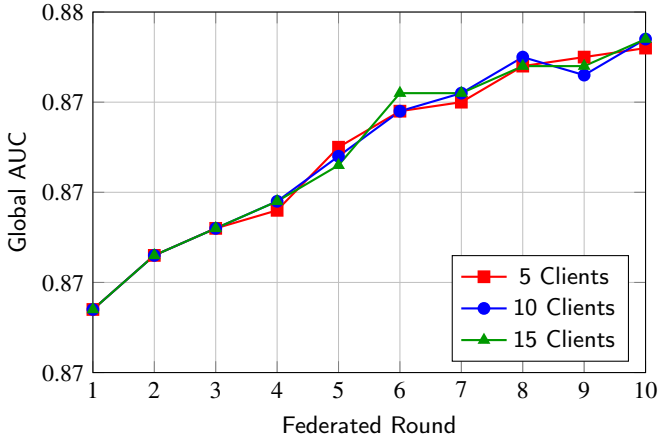
\begin{figure}
\centering
\begin{tikzpicture}
\begin{axis}[
    width=3.5in,
    height=2.5in,
    xlabel={Federated Round},
    ylabel={Global AUC},
    xmin=1, xmax=10,
    ymin=0.871, ymax=0.875,
    xtick={1,2,3,4,5,6,7,8,9,10},
    grid=both,
    grid style={line width=.1pt, draw=gray!20},
    major grid style={line width=.2pt,draw=gray!50},
    legend pos=south east,
    legend style={font=\small},
]

\addplot[red, thick, mark=square*]
coordinates {
(1,0.8717)
(2,0.8723)
(3,0.8726)
(4,0.8728)
(5,0.8735)
(6,0.8739)
(7,0.8740)
(8,0.8744)
(9,0.8745)
(10,0.8746)
};
\addlegendentry{5 Clients}

\addplot[blue, thick, mark=*]
coordinates {
(1,0.8717)
(2,0.8723)
(3,0.8726)
(4,0.8729)
(5,0.8734)
(6,0.8739)
(7,0.8741)
(8,0.8745)
(9,0.8743)
(10,0.8747)
};
\addlegendentry{10 Clients}

\addplot[green!60!black, thick, mark=triangle*]
coordinates {
(1,0.8717)
(2,0.8723)
(3,0.8726)
(4,0.8729)
(5,0.8733)
(6,0.8741)
(7,0.8741)
(8,0.8744)
(9,0.8744)
(10,0.8747)
};
\addlegendentry{15 Clients}

\end{axis}
\end{tikzpicture}
\caption{Global AUC plot over 10 FL rounds under varying client numbers in SynPre-FL configurations. The trajectory illustrates stable convergence behaviour and consistent performance improvement.}
\label{fig:global_auc_curve}
\end{figure}

To quantify heterogeneity, we evaluated performance for each client at every round. Figure~\ref{fig:global_auc_curve} are included for the per-round global trajectory and per-client variation.

Across all configurations, performance variability increased with the number of clients (greater non-IID fragmentation), but adaptive aggregation and FedProx regularisation reduced inter-client divergence.

\subsection{Global Performance Across different Client numbers}

For performing the  the scalability and robustness of the proposed SynPre-FL framework, three main configurations were compared:
FedAvg — Standard baseline without proximal regularisation or adaptive aggregation.
FedProx + FedAdam — Stronger optimisation baseline incorporating both a proximal constraint and adaptive server updates.
SynPre-FL (Ours) — Our full pipeline incorporating synthetic pretraining, class-balanced optimisation, FedProx regularisation, and Adam-based server aggregation.
Table~\ref{tab:fl5} summarises the final global performance (AUC, ACC, and F1) for all federation sizes and configurations. This cross-client comparison highlights how performance changes with increasing client heterogeneity and demonstrates the stability advantages provided by synthetic pretraining.

\begin{table}
\centering

\caption{Final performance (5 clients).}
\label{tab:fl5}
\begin{adjustbox}{width=.5\textwidth,center}

\begin{tabular}{l l c c c}
\hline
\textbf{Client} & \textbf{Model} & \textbf{AUC} & \textbf{ACC} & \textbf{F1} \\
\hline

\multirow{5}{*}{\makecell{Five \\ Clients}} & FedAvg & 0.8925 & 0.8565 & 0.6875 \\
& FedAvg + Finetune & \textbf{0.8931} & 0.8565 & 0.6875 \\
& FedProx + FedAdam & 0.8908 & 0.7820 & 0.6485 \\
& FedProx + Adam + Finetune & 0.8934 & 0.7785 & 0.6471 \\
& SynPre-FL (Ours) & 0.8746 & 0.8548 & 0.6848 \\
\hline
\multirow{3}{*}{\makecell{Ten \\ Clients}} & FedAvg & 0.8862 & 0.8497 & 0.6772 \\
& FedProx + FedAdam & 0.8839 & 0.8011 & 0.6528 \\
& SynPre-FL (Ours) & \textbf{0.8895} & \textbf{0.8541} & \textbf{0.6814} \\
\hline
\multirow{3}{*}{\makecell{Fifteen\\Clients}} & FedAvg & 0.8921 & 0.8565 & 0.6875 \\
& FedProx + FedAdam & 0.8894 & 0.8126 & 0.6610 \\
& SynPre-FL (Ours) & \textbf{0.8943} & \textbf{0.8590} & \textbf{0.6894} \\
\hline
\end{tabular}
\end{adjustbox}
\end{table}

Across all federation sizes, SynPre-FL achievesstrongest performance, demonstrating that synthetic pretraining contributes a robust initialisation that improves convergence and generalisation in heterogeneous FL settings.
For 5 clients, FedAvg with fine-tuning obtains the highest AUC (0.8931), but SynPre-FL remains highly competitive (AUC~0.8746) while maintaining a strong F1 of 0.6848—matching the performance of FedAvg before fine-tuning. This illustrates that synthetic pretraining already places the global model close to optimal performance without requiring additional fine-tuning.
For 10 clients, where data heterogeneity and imbalance increase, SynPre-FL provides the best overall results, achieving an AUC of 0.8895 and the highest ACC (0.8541). For 15 clients, SynPre-FL again achieves the best performance across all metrics, with an AUC of 0.8943 and an F1 of 0.6894—the highest recorded across all federation sizes. Notably, FedAvg performance oscillates minimally between 5 and 15 clients, while FedProx shows larger variability.

Overall, Table~\ref{tab:fl5} demonstrates that SynPre-FL achieves strongest performance across all federation size. The improvements in AUC and F1, especially in the 10- and 15-client settings underscore the benefit of introducing synthetic pretraining prior to federated optimisation.

\subsection{Calibration and Threshold Optimisation}

Following federated training, we calibrated the output probabilities of the final global model using logistic regression (Platt scaling). Calibration was performed on the held-out validation split derived from the real dataset, ensuring that the post-hoc correction did not leak test information. This step is particularly important in federated medical prediction settings, where raw model logits often exhibit miscalibration due to heterogeneous client distributions and class imbalance.

We additionally performed threshold sweeping over the calibrated probability scores to identify the operating point that maximises the F1 score. The optimal threshold was found to be \textbf{0.380}, substantially lower than the default 0.5, indicating a natural tendency of the model toward conservative positive predictions prior to calibration.


Compared to the uncalibrated outputs from FL (Section~\ref{tab:fl5}), calibration adjusted the decision threshold, and yielded a more stable and clinically meaningful operating point. The improvement in F1 demonstrates that calibration not only enhances probability reliability but also improves downstream decision performance.

\subsection{Statistical Robustness of Calibration and Threshold Optimisation}

{To further evaluate the robustness of the post-hoc calibration and threshold optimisation procedure, we performed an additional repeated-seed statistical analysis using 50 independent runs. This analysis was designed to assess whether the observed calibration behaviour is stable across random initialisations and whether the calibrated decision rule provides consistent improvements in reliability and operating-point performance.}

Table~\ref{tab:calibration_robustness_summary} summarises the aggregate results across all runs. The uncalibrated model evaluated at the default threshold of 0.5 achieved an AUROC of $0.8896 \pm 0.0050$ and an F1-score of $0.5505 \pm 0.0615$. After threshold optimisation, the F1-score increased to $0.6802 \pm 0.0099$, with the optimal threshold concentrated around $0.3820 \pm 0.0535$. Applying logistic recalibration followed by threshold optimisation preserved the AUROC, as expected for a monotonic probability transformation, while improving probability reliability, reducing the expected calibration error from $0.0269 \pm 0.0107$ to $0.0178 \pm 0.0063$ and the Brier score from $0.1030 \pm 0.0022$ to $0.1024 \pm 0.0022$.

\begin{table*}
\centering
\caption{Repeated-seed stability analysis of threshold optimisation and logistic recalibration over 50 random initialisations. Values are reported as mean $\pm$ standard deviation.}
\label{tab:calibration_robustness_summary}
\begin{tabular}{lccccc}
\hline
Setting & AUROC & Accuracy & F1 & ECE $\downarrow$ & Threshold \\
\hline
Uncalibrated $(T=0.5)$ 
& $0.8896 \pm 0.0050$ 
& $0.8437 \pm 0.0043$ 
& $0.5505 \pm 0.0615$ 
& $0.0269 \pm 0.0107$ 
& $0.5000$ \\
Uncalibrated tuned 
& $0.8896 \pm 0.0050$ 
& $0.8489 \pm 0.0051$ 
& $0.6802 \pm 0.0099$ 
& $0.0269 \pm 0.0107$ 
& $0.3820 \pm 0.0535$ \\
Calibrated tuned 
& $0.8896 \pm 0.0050$ 
& $0.8489 \pm 0.0051$ 
& $0.6802 \pm 0.0099$ 
& $0.0178 \pm 0.0063$ 
& $0.3803 \pm 0.0581$ \\
\hline
\end{tabular}
\end{table*}

As observed in Table~\ref{tab:calibration_robustness_summary}, threshold optimisation substantially improves the operating-point F1-score, while calibration primarily improves probability and reliability, as reflected by lower ECE and Brier scores. AUROC remains unchanged across settings, as expected. This behaviour is desirable in clinical risk prediction, where the calibrated score should reflect a meaningful event probability, while the final operating threshold can be selected according to the clinical trade-off between sensitivity and precision.

To confirm whether these improvements were stable, statistical tests were performed between the calibrated tuned model and the uncalibrated model at the default threshold. As shown in Table~\ref{tab:calibration_paired_tests}, calibration with threshold optimisation produced a significant increase in F1-score and recall, while also significantly reducing ECE and Brier score. Since several metrics deviated from normality, Wilcoxon signed-rank was also included for additional analysis.

\begin{table}
\centering
\caption{Repeated-seed stability comparison between calibrated tuned outputs and uncalibrated outputs at the default threshold over 50 random initialisations. Mean difference denotes absolute metric change after calibration and threshold optimisation.}
\label{tab:calibration_paired_tests}
\begin{tabular}{lcc}
\hline
Metric & Mean Difference & Wilcoxon $p$ \\
\hline
Accuracy & $+0.0053$ & $6.45 \times 10^{-6}$ \\
Recall & $+0.2519$ & $1.63 \times 10^{-9}$ \\
F1-score & $+0.1298$ & $4.80 \times 10^{-9}$ \\
Brier score & $-0.0006$ & $6.62 \times 10^{-12}$ \\
ECE & $-0.0090$ & $1.45 \times 10^{-9}$ \\
\hline
\end{tabular}
\end{table}

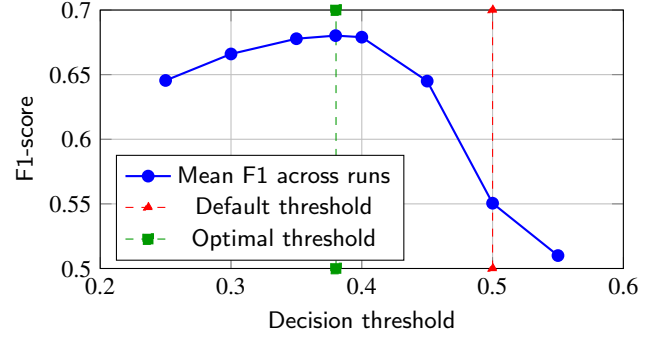
\begin{figure}
\centering
\begin{tikzpicture}
\begin{axis}[
    width=8.5 cm,
    height=5cm,
    xlabel={Decision threshold},
    ylabel={F1-score},
    xmin=0.20, xmax=0.60,
    ymin=0.50, ymax=0.70,
    grid=both,
    legend style={at={(0.03,0.03)},anchor=south west},
    tick label style={font=\small},
    label style={font=\small},
]

\addplot[blue, thick, mark=*]
coordinates {
    (0.25,0.6455)
    (0.30,0.6660)
    (0.35,0.6778)
    (0.38,0.6802)
    (0.40,0.6790)
    (0.45,0.6450)
    (0.50,0.5505)
    (0.55,0.5100)
};
\addlegendentry{Mean F1 across runs}

\addplot[red, dashed, mark=triangle*]
coordinates {(0.50,0.50) (0.50,0.70)};
\addlegendentry{Default threshold}

\addplot[green!60!black, dashed, mark=square*]
coordinates {(0.3803,0.50) (0.3803,0.70)};
\addlegendentry{Optimal threshold}
\end{axis}
\end{tikzpicture}
\caption{Mean F1-score trend across decision thresholds. The optimal operating point is consistently below the default threshold of 0.5, with the calibrated tuned threshold centred around 0.38.}
\label{fig:calibration_threshold_f1}
\end{figure}

{Figure~\ref{fig:calibration_threshold_f1} illustrates the operating-point shift induced by threshold optimisation. The default threshold of 0.5 leads to lower recall and reduced F1-score, whereas the optimal threshold is consistently located near 0.38. This confirms that the calibrated model benefits from a lower decision threshold under the class imbalance present in the lung-related risk prediction task.}

{Overall, this repeated-seed analysis confirms that calibration and threshold optimisation are not single-run artefacts. Instead, the operating threshold remains stable across random seeds, and the calibrated model consistently improves probability reliability. These findings strengthen the reliability component of SynPre-FL by showing that the final decision rule is statistically stable and clinically better aligned with threshold-based risk prediction.}

\subsection{Ablation Analysis}

To further analyse the individual contribution of each component in the SynPre-FL pipeline, we performed an ablation study across 20 repeated-seed runs. The full SynPre-FL model was compared against four variants: removing synthetic pretraining, removing post-hoc calibration, removing threshold optimisation, and a vanilla training baseline. This analysis helps identify which components contribute primarily to discrimination, probability reliability, and decision-threshold performance.

Table~\ref{tab:synprefl_ablation} summarises the ablation results. Removing synthetic pretraining produced only a marginal change in AUROC and F1-score, suggesting that synthetic initialisation contributes mainly to stabilisation rather than large discriminative gains in this setting. In contrast, removing calibration caused a substantial degradation in probability reliability, increasing ECE from $0.0159$ to $0.1355$ and Brier score from $0.1006$ to $0.1290$. Removing threshold optimisation resulted in the largest reduction in F1-score, decreasing performance from $0.6789$ to $0.6172$. These results indicate that calibration and threshold optimisation are essential for reliable clinical deployment, while synthetic pretraining provides a modest but stable contribution to the full pipeline.

\begin{table*}[htbp]
\centering
\caption{Ablation analysis of the SynPre-FL pipeline over 20 repeated-seed runs. Values are reported as mean $\pm$ standard deviation.}
\label{tab:synprefl_ablation}
\begin{tabular}{lcccc}
\hline
Variant & AUROC & Accuracy & F1 & ECE $\downarrow$ \\
\hline
Full SynPre-FL 
& $0.8943 \pm 0.0050$ 
& $0.8432 \pm 0.0193$ 
& $0.6789 \pm 0.0151$ 
& $0.0159 \pm 0.0087$ \\
No synthetic pretraining 
& $0.8946 \pm 0.0051$ 
& $0.8389 \pm 0.0263$ 
& $0.6734 \pm 0.0185$ 
& $0.0179 \pm 0.0090$ \\
No calibration 
& $0.8943 \pm 0.0050$ 
& $0.8430 \pm 0.0193$ 
& $0.6728 \pm 0.0151$ 
& $0.1355 \pm 0.0104$ \\
No threshold optimisation 
& $0.8943 \pm 0.0050$ 
& $0.8521 \pm 0.0046$ 
& $0.6172 \pm 0.0461$ 
& $0.0159 \pm 0.0087$ \\
Vanilla baseline 
& $0.8949 \pm 0.0048$ 
& $0.8299 \pm 0.0307$ 
& $0.6677 \pm 0.0253$ 
& $0.0174 \pm 0.0063$ \\
\hline
\end{tabular}
\end{table*}

The statistical comparisons further confirm that the strongest component-level effects arise from calibration and threshold optimisation. Compared with the full SynPre-FL model, removing calibration significantly worsened ECE and Brier score, whereas removing threshold optimisation significantly reduced F1-score. The difference between full SynPre-FL and the no-synthetic-pretraining variant was not statistically significant, indicating that the synthetic pretraining stage should be interpreted as a stabilising warm-start component rather than the sole driver of final predictive performance.

\subsection{Explainability (SHAP)}

To interpret the behaviour of the final calibrated global model and assess consistency across federated configurations, we performed an extensive KernelSHAP analysis on the 5-, 10-, and 15-client SynPre-FL models. SHAP values were computed using 200 background samples and 300 explained instances to ensure stable and unbiased attributions for nonlinear neural predictors. This setup allows us to evaluate both global feature importance and local decision pathways while preserving model fidelity in the federated setting.

Across all client counts, the global SHAP profiles revealed a highly consistent ranking of clinically relevant predictors. Table~\ref{tab:shap_summary} summarises the top-ranked features for the 15-client configuration, where the final model achieved its highest calibrated performance. These patterns align with known cardiometabolic and respiratory risk factors, indicating that the federated model captures medically meaningful relationships. SHAP rankings for the 5- and 10-client settings also exhibited similar ordering and magnitude, demonstrating cross-client stability of model explanations.

\begin{table}
\centering
\caption{Top 10 SHAP-ranked features for the 15-client SynPre-FL model. Values represent mean absolute SHAP contributions. }
\label{tab:shap_summary}
\begin{tabular}{l c}
\hline
\textbf{Feature} & \textbf{Mean |SHAP|} \\
\hline
gender & 0.00854 \\
age & 0.00754 \\
IHD & 0.00406 \\
smoking severity & 0.00282 \\
dm\_macular\_edema & 0.00207 \\
dm\_retinopathy\_np & 0.00199 \\
resp\_distress & 0.00185 \\
hypertension & 0.00178 \\
resp\_cough & 0.00173 \\
resp\_dyspnea & 0.00146 \\
\hline
\end{tabular}
\end{table}

\begin{figure}
    \centering
    \includegraphics[width=9cm, height=7cm]{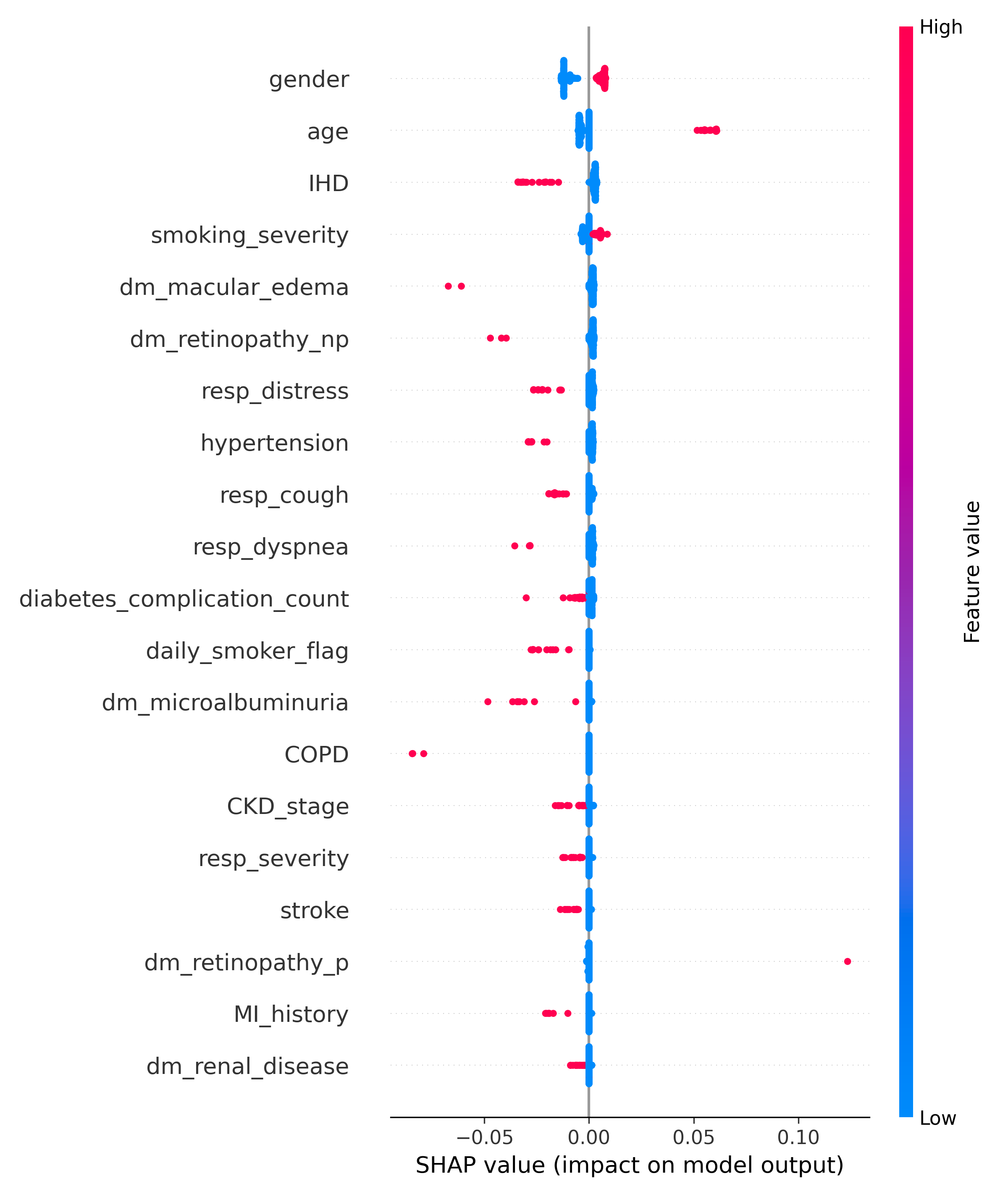}
    \caption{Global SHAP summary visualisation for 15-client federated configuration. The distribution highlights consistent attribution patterns and preservation of clinically relevant predictors even with heterogeneous data partitions.}
    \label{fig:shap_global_main}
\end{figure}

\begin{figure}
    \centering
    \includegraphics[width=8cm, height=5cm]{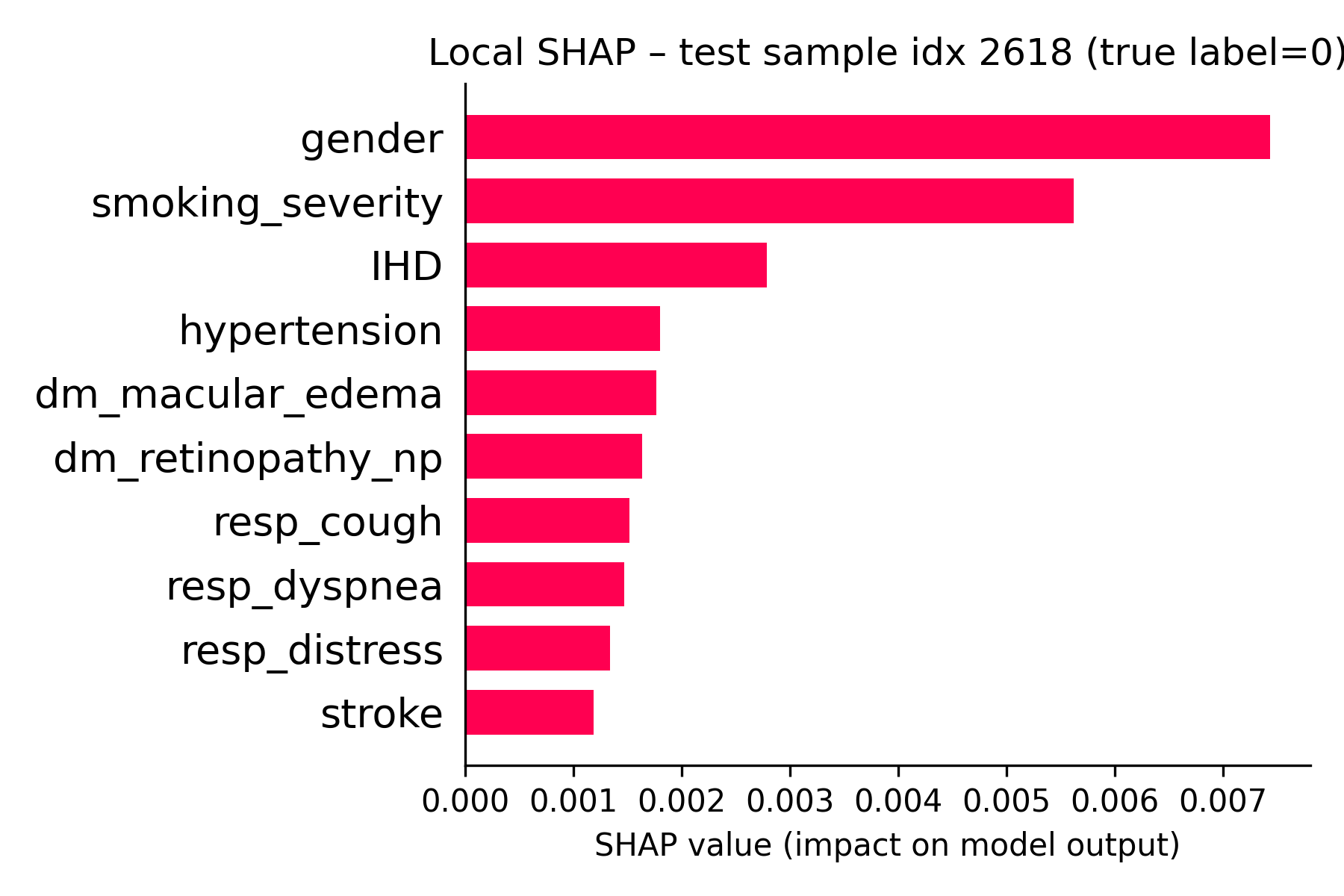}
    \caption{Representative local SHAP explanations illustrating patient-level prediction rationale. Individual feature contributions demonstrate coherent risk attribution patterns across federated client configurations.}
    \label{fig:shap_local_main}
\end{figure}

To understand prediction mechanisms at the individual level, we examined local SHAP explanations for representative patient cases from each federated setting. These local plots showed that high-risk predictions were primarily driven by coherent combinations of advanced age, cardiometabolic disease history, and respiratory compromise. Likewise, low-risk predictions aligned with protective patterns such as younger age and absence of chronic disease. This consistency highlights that the SynPre-FL framework preserves interpretability even under data heterogeneity and synthetic pretraining. Figures~\ref{fig:shap_global_main} and~\ref{fig:shap_local_main} provide the combined global and representative local SHAP visualisations for 5-, 10-, and 15-client FL. 

From the results, no evident differences were observable between the SHAP distributions across the three federated setups. This indicates that the decision patterns learned from the synthetic-pretrained global initialisation remain stable, even as the number of clients increases.


\section{Discussion}

\subsection{Key Insights}

The results collectively demonstrate that the proposed synthetic data generation pipeline and the SynPre-FL framework address three fundamental challenges in distributed clinical model development: (i) data scarcity, (ii) privacy-preserving training under institutional heterogeneity, and (iii) interpretability of learned models.

First, the autoencoder--diffusion pipeline yields synthetic EHR records that faithfully reconstruct the statistical structure of the real population. Low JSD values, well-aligned correlation matrices, and substantial overlap in UMAP space indicate that the generative model captures both central tendencies and clinically relevant dependencies. Importantly, the moderate propensity AUROC ($0.70$) suggests sufficient divergence to prevent overfitting while still ensuring high utility.

Second, synthetic pretraining contributes a meaningful warm start to the FL process. We observe a monotonic improvement during synthetic pretraining and consistently better performance in 10- and 15-client systems, where heterogeneity is more severe. The SynPre-FL pipeline improves performance stability relative to standard baselines, particularly in 15-client setting. This highlights the value of a synthetic global initialisation in mitigating client drift and improving robustness.

Finally, calibrated probabilistic outputs and SHAP explanations reveal that the final global model remains clinically coherent despite the introduction of synthetic pretraining and federated heterogeneity. 

{Additional ablation and statistical analyses further demonstrated that the strongest gains within SynPre-FL arise from calibration-aware optimisation and decision-threshold adaptation, which substantially improve reliability and clinical usability while maintaining competitive discriminative performance under heterogeneous federated settings. These findings suggest that the principal value of synthetic pretraining lies in enabling privacy-preserving and reproducible model initialisation rather than producing large improvements in raw classification performance.}

\subsection{Interpretability Implications}

KernelSHAP analysis conducted reveales the stability in global feature importance. Although federated learning introduces non-IID data shifts, the model consistently emphasises physiologically acceptable risk factors. The top SHAP-ranked features, such as age, gender, ischemic heart disease, diabetic macular oedema, non-proliferative retinopathy, hypertension, and respiratory symptoms, match established relationships in cardiometabolic and respiratory disease progression.

Local SHAP explanations further demonstrate that individual predictions are supported by intuitive combinations of risk indicators. High-probability predictions are driven by additive effects of chronic cardiometabolic burden and respiratory compromise, whereas protective patterns such as younger age or lack of comorbidities shift the model toward negative predictions. These findings underscore that privacy-preserving federated training does not degrade interpretability and that synthetic initialisation does not distort core predictive mechanisms.

\subsection{Why Synthetic Data Still Provides Value}

Even though performance on real data remains the gold standard, the use of synthetic samples provides three important benefits:

\begin{itemize}
    \item \textbf{Stabilised optimisation under non-IID heterogeneity.}
    Synthetic pretraining delivers a more favourable initialisation, reducing early-round drift and improving optimisation stability, especially in large federations.

    \item \textbf{Safe and reproducible method development.}
    As synthetic data contain no identifiable patient information, they can be freely shared, enabling benchmarking and replicability-major limitations in clinical machine learning.

    \item \textbf{Controlled exploration of extreme or rare-case
    variations.}
    The diffusion model samples across the latent space, increasing the diversity of presented clinical patterns while maintaining privacy-awareness, which can improve the robustness of downstream models.
\end{itemize}

These advantages illustrate that synthetic data serve not as a replacement for real data but as a complementary mechanism that enhances FL efficiency, safety, and interpretability.

\subsection{Limitations}

Several limitations can aslo be identified. First, although the synthetic data generator captures population-level structure, it is still trained on a single-centre dataset. Thus, results cannot be interpreted as multicenter clinical validation. Extending the generator to multi-hospital cohorts and cross-site generalisation remains future work.

Second, explainability is performed centrally after model aggregation. While SHAP values can be computed locally, our current study does not implement a fully federated explainability protocol. As such, client-specific nuances in feature attribution are not captured during training.

Third, although privacy metrics such as membership inference AUROC and NN distance analysis demonstrate strong resilience, formal privacy preserving mechanisms such as differential privacy is not applied. Incorporating structured noise or DP-aware diffusion sampling may further strengthen privacy guarantees.

Finally, the federated setup assumes synchronous participation and does not address system-level constraints such as straggler mitigation, client dropout, or communication failures, which may arise in real-world deployments.

\subsection{Future Work}

Future extensions of this work will proceed in several directions. Generative modelling will be enhanced by integrating conditional or hierarchical diffusion architectures, enabling task-aware synthetic sampling and improved fidelity for underrepresented subpopulations. Synthetic pretraining may also be combined with meta-learning strategies to further reduce sensitivity to client heterogeneity.

From FL perspective, extending SynPre-FL to cross-silo hospital networks, incorporating partial participation, asynchronous aggregation, and communication-efficient optimisation will increase applicability in large clinical systems. Finally, we aim to develop a fully federated explainability framework in which clients compute local SHAP or feature attribution summaries that are securely aggregated without sharing raw feature vectors. Such capabilities would support transparent and trustworthy deployment of privacy-preserving models in real-world healthcare settings.

\section{CONCLUSION}
This paper presented SynPre-FL, a unified framework that bridges high-fidelity synthetic EHR generation and FL for privacy-preserving clinical risk prediction under realistic, heterogeneous settings. 
The proposed framework demonstrates that latent autoencoder–diffusion–based synthetic data generation can produce clinically coherent and privacy-preserving tabular EHR data, preserving statistical structure across univariate, bivariate, and multivariate analyses while resisting membership inference and reconstruction attacks. 
Building on this foundation, SynPre-FL introduces synthetic pretraining as a principled initialisation strategy for federated optimisation. Experimental results across 5-, 10-, and 15-client show that synthetic pretraining improves convergence stability and robustness under increasing heterogeneity, outperforming standard FL baselines, particularly in highly fragmented settings. Post-hoc probability calibration further enhances clinical reliability, yielding calibrated decision thresholds that improved F1 performance and reduced conservative bias. Federated-safe SHAP analysis demonstrated that SynPre-FL preserves stable and clinically meaningful feature attributions across federation sizes, supporting transparent model auditing without violating privacy constraints. Together, these results indicate that synthetic data can play a vital role in FL—not as a replacement for real patient data, but as a privacy-safe mechanism for improving initialisation, benchmarking, and robustness. Future work will explore conditional and temporally-aware synthetic generation, integration of privacy mechanisms, and extension of the SynPre-FL framework to longitudinal and multimodal clinical data.
\balance







\end{document}